\documentclass[10pt,twocolumn,letterpaper]{article}

\usepackage{cvpr}
\usepackage{times}
\usepackage{epsfig}
\usepackage{graphicx}
\usepackage{amsmath}
\usepackage{amssymb}
\usepackage{algorithm}
\usepackage{color}
\usepackage{algorithmic}
\usepackage{subfig}
\usepackage{booktabs}
\usepackage{authblk}
\newcommand{\J}[1]{{#1}} 

\usepackage[breaklinks=true,bookmarks=false]{hyperref}

\cvprfinalcopy 

\def\cvprPaperID{****} 

\ifcvprfinal\fi
\begin{document}

\title{Deep Self-Taught Learning for Weakly Supervised Object Localization}


\author{Zequn Jie$^{*\dag}$ \quad Yunchao Wei$^*$ \quad Xiaojie Jin$^*$ \quad Jiashi Feng$^*$ \quad Wei Liu$^\dag$ \\
	$^*$National University of Singapore \quad  $^\dag$Tencent AI Lab \\ {\tt\small \{elejiez, eleweiyv, elefjia\}@nus.edu.sg \quad xiaojie.jin@u.nus.edu \quad wliu@ee.columbia.edu}
}
\maketitle
\thispagestyle{empty}

\begin{abstract}
	
Most existing weakly supervised localization (WSL) approaches learn detectors by finding positive bounding boxes based on features learned with image-level supervision. However, those features do not contain spatial location related information and usually provide poor-quality positive samples for training a detector. To overcome this issue, we propose a deep self-taught learning approach, which makes the detector learn the object-level features reliable for acquiring tight positive samples and afterwards re-train itself based on them. Consequently, the detector progressively improves its detection ability and localizes more informative positive samples. To implement such self-taught learning, we propose a seed sample acquisition method via image-to-object transferring and dense subgraph discovery to find reliable positive samples for initializing the detector. An online supportive sample harvesting scheme is further proposed to dynamically select the most confident tight positive samples and train the detector in a mutual boosting way. To prevent the detector from being trapped in poor optima due to overfitting, we propose a new relative improvement of predicted CNN scores for guiding the self-taught learning process. Extensive experiments on PASCAL 2007 and 2012 show that our approach outperforms the state-of-the-arts, strongly validating its effectiveness.
\end{abstract}

\vspace{-0.5cm}
\section{Introduction}


Weakly Supervised Localization (WSL) refers to learning to localize objects within images with only image-level annotations that simply indicate the presence of an object category. WSL is gaining increasing importance in large-scale vision applications because it does not require expensive bounding box annotations like its fully-supervised counterpart~\cite{sermanet2013overfeat,girshick2014rich,girshick2015fast,ren2015faster,liang2016reversible,jie2016tree} in the model training phase.
\begin{figure}[t]
	\begin{center}
		\includegraphics[width=1\linewidth, height=0.8\linewidth]{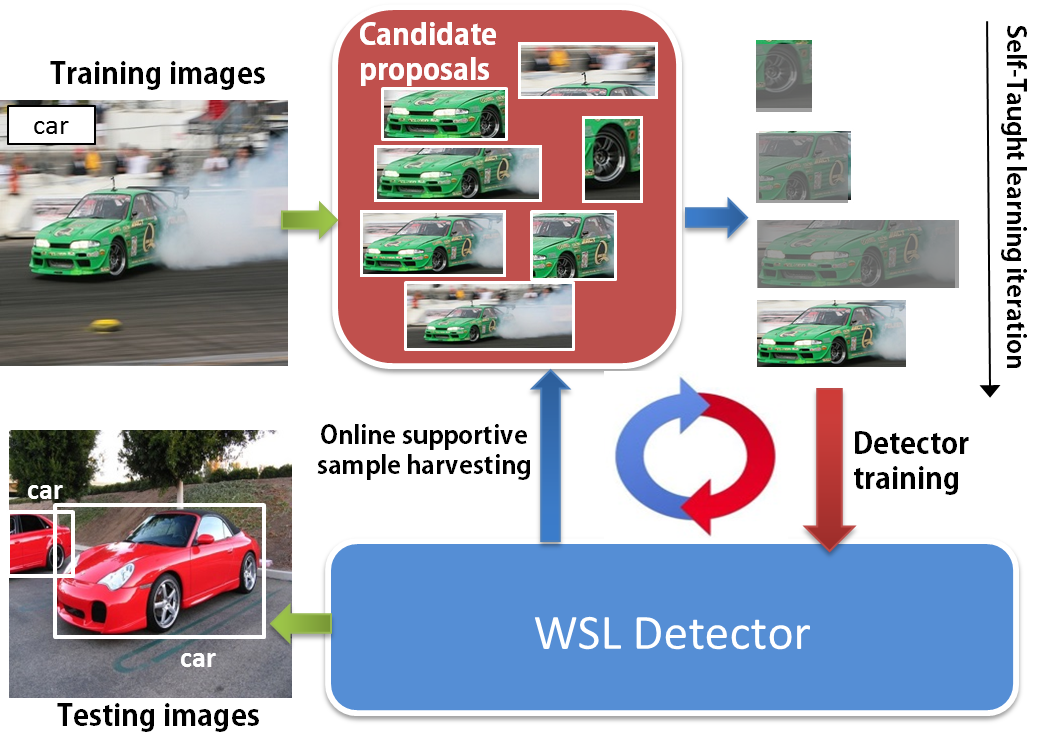}
	\end{center}
	\vspace{-0.4cm}
	\caption{ An illustration of deep self-taught learning for weakly supervised object localization. Given image-level supervision, seed positive proposals are first obtained as initial positive samples for a CNN detector. The CNN detector is then trained with self-taught learning which alternates between training and online supportive sample harvesting relying on the \emph{relative improvement} of CNN scores predicted by the detector. }
	\label{fig:illus}
	\vspace{-0.5cm}
\end{figure}

WSL is a challenging problem due to the insufficiency of information for learning a good detector. Correctly identifying the reliable positive samples (bounding boxes) from a collection of candidates is thus of critical importance. Most previous WSL methods \cite{bilen2014weakly,cinbis2015multifold,li2016domain,siva2011weakly}  discover high-confident positive samples from the images with positive annotations by applying multiple instance learning (MIL) or other similar algorithms.   Recent WSL methods \cite{hoffman2014lsda,hoffman2015detector,rochan2015weakly,shi2012transfer,li2016domain} also combine deep convolutional neural network (CNN) models \cite{krizhevsky2012imagenet,szegedy2015going,he2016deep} with MIL, considering that CNN architectures can provide more powerful image representations. However, the representation provided by a CNN tailored to classification does not contain any specific information about object spatial locations and is thus not suitable for object-level localization tasks, leading to marginal benefits for learning a high-quality object detector.

Moreover, such methods only perform off-line MIL to mine confident class-specific object proposals before training the detector, where the strong discriminating power of the learned object-level CNN detector is not fully leveraged to mine high-quality proposals for  detector learning.


In this paper, we propose to make a weak detector ``train'' itself through exploiting a novel deep self-taught learning approach such that it progressively gains a stronger ability for object detection and solves the WSL problem, as illustrated in Fig.~\ref{fig:illus}. This is a new WSL paradigm and can address the above issues of the existing methods.


Given several seed positive proposals, self-taught learning enables the detector to spontaneously harvest the most confident tight positive proposals (called supportive samples) in an online manner, through examining their predicted scores from the detector itself. By fully exploiting the strong discriminating ability of the regional CNN detector (\emph{e.g.}, Fast R-CNN \cite{girshick2015fast}),  supportive samples of higher quality can be identified, compared with the ones provided by the conventional CNN plus MIL approaches. However, one key problem with the above online supportive sample harvesting strategy for self-taught learning is that some poor seed positive samples may be easily fitted by the CNN detector due to its strong learning ability and hence trap the CNN detector in poor local optima. To address this critical problem pertaining to self-taught learning, we propose a novel \emph{relative improvement} metric for facilitating supportive sample harvesting. The {relative improvement} of scores can effectively filter those suspicious samples whose high predicted scores are from undesired overfitting, thereby helping identify authentic samples of high-quality.


The very first step of the above self-taught learning process is to acquire high-quality seed positive samples. We propose an image-to-object transferring scheme to find reliable seed positive samples. Concretely, we first select the object proposals with high responses\footnote[1]{Throughout this paper, response and CNN score refer to the final probability output after softmax normalization to the target class.} to the target class obtained by training a multi-label classification network. Selecting samples in this way roughly establishes a correspondence between image-level annotations and object-level high-response proposals. Then we propose to employ a dense subgraph discovery method to select a few dense spatially distributed proposals as the seed positive samples, by exploiting the spatial correlations for selected proposals as above. Comprehensive experiments demonstrate the effectiveness of our proposed approach for acquiring reliable seed samples, and the obtained seed samples are indeed beneficial for the following self-taught learning procedure to tackle WSL problems.

To sum up, we make the following contributions to WSL in this work: 
\begin{enumerate}
	\setlength\itemsep{0em}
	\item We propose a novel deep self-taught learning approach to progressively harvest high-quality positive samples guided by the detector itself, therefore significantly improving the quality of positive samples during detector training. 
	\item A novel relative score improvement based selection strategy is proposed to prevent the detector from being trapped in poor local optima resulting from the overfitting to seed positive samples.
	\item  To acquire high-quality seed positives, we propose a novel image-to-object transferring technique to learn the spatial-aware features tailored to WSL. To further incorporate the spatial correlations between the selected object samples, a novel dense subgraph discovery based method is proposed to mine the most confident class-specific samples from a set of  spatially highly correlated candidate samples.
\end{enumerate}

\vspace{-0.3cm}
\section{Related Work}
Previous works on WSL can be roughly categorized into  MIL based methods and end-to-end CNN models.

Actually, the majority of existing methods formulate WSL as an MIL problem. Given weak image-level supervisory information, these methods typically alternate between learning a discriminative representation of the object and selecting the positive object samples in positive images based on this representation. However, this results in a non-convex optimization problem, so these methods are prone to being trapped in local optima, and their solutions are sensitive to the initial positive samples. Many efforts have been made to address the above issue. Deselaers \emph{et al.}~\cite{deselaers2010localizing} initialized object locations using the objectness method~\cite{alexe2012measuring}. Siva \emph{et al.}~\cite{siva2012defence} selected positive samples by maximizing the distances between the positive samples and those in negative images. Bilen \emph{et al.}~\cite{bilen2014weakly} proposed a smoothed version of MIL that softly labels object proposals instead of choosing the highest scoring ones. Song \emph{et al.}~\cite{song2014learning} proposed a graph-based method to initialize the object locations by solving a submodular cover problem. Wang \emph{et al.}~\cite{wang2014weakly} proposed a latent semantic clustering method to select the most discriminative cluster for each class based on Probability Latent Semantic Analysis (pLSA).

Apart from improving the initial quality of positive samples, some work also focuses on improving optimization during iterative training. Singh \emph{et al.}~\cite{singh2012unsupervised} iteratively trained SVM classifiers on a subset of the initial positive samples, and evaluated them on another set to update the training samples. Bilen \emph{et al.}~\cite{bilen2014weakly} proposed a posterior regularization formulation that regularizes the latent (object location) space by penalizing unlikely configurations based on symmetry and mutual exclusion of objects. Cinbis \emph{et al.}~\cite{cinbis2015multifold} proposed a multi-fold training strategy to alleviate the local optimum issue.

End-to-end CNN models are also used for WSL. Bilen \emph{et al.}~\cite{bilen2016weakly} proposed an end-to-end CNN model with two streams, one for classification and the other for localization, which outputs final scores for the proposals by the element-wise multiplication on the results of the two streams. Kantorov \emph{et al.}~\cite{kantorov2016contextlocnet} proposed a context-aware CNN model trained with contrast-based contextual guidance, resulting in refined boundaries of detected objects.

Perhaps \cite{li2016domain} is the closest work to ours.  \cite{li2016domain} first trains a whole-image multi-label classification network and then selects confident class-specific proposals with a mask-out strategy and MIL. Finally, a Fast R-CNN detector is trained on these proposals. However, the whole-image classification in \cite{li2016domain} may not provide suitable features for object localization which requires tight spatial coverage of the whole object instance. Additionally, SVM is used in MIL in \cite{li2016domain}, which has the inferior discriminating ability to the regional CNN detector. In contrast, our approach overcomes this weakness by performing image-to-object transferring during multi-label image classification and online supportive sample harvesting in regional CNN detector learning.

\section{Deep Self-Taught Learning for WSL}
In this section, the proposed deep self-taught learning approach for WSL will be detailed. We first describe the image-to-object transferring and dense subgraph discovery based methods used to acquire high-quality seed positive samples for detector self-taught learning. Then, online supportive sample harvesting is presented, which progressively improves the quality of the positive samples, where the detector dynamically harvests the most informative positive samples during learning, guided by the relative CNN score improvement from the detector itself.

\subsection{Seed Sample Acquisition}
\subsubsection{Image-to-Object Transfer}
We propose an image-to-object transferring approach to identify reliable seed samples with highest class-specific likelihood, given only image-level annotations. Considering that each positive image contains at least one positive object proposal that contributes significantly to each class, we train a multi-label classification CNN model as the first step to identify seed samples.  We follow the  method Hypothesis-CNN-Pooling (HCP)~\cite{wei2015hcp} in multi-label classification to mine the proposals which contribute most to image-level classification. Specifically, HCP accepts a number of input proposals and feeds them into the CNN classification network. Then cross-proposal max-pooling is performed in the integrative prediction stage for each class.

More formally,  assume that ${\{\mathbf{v_{i}}\}}_{i=1}^{n}$  is the output response vector of the $i$-th proposal from the CNN, and that $\{v_{i}^{j}\}_{j=1}^{c}$ is the output response of the $j$-th class in $\mathbf{v_{i}}$. The final integrative prediction for an image on the $j$-th class is
\begin{equation*}
	v^{j}= \max(v_{1}^{j}, v_{2}^{j},\ldots,v_{n}^{j}).
\end{equation*}
With cross-proposal max-pooling, the highest predicted response corresponding to the object of the target class will be reserved, while the responses from the negative objects will be ignored. In this way, the image-level classification error will only be back-propagated through the most confident proposal such that the network achieves spatial-awareness during training. This fills the gap between the image-level annotation and the object-level features, thus providing more discriminative features for the object-level detection task. More  details of HCP can be found in \cite{wei2015hcp}.

\begin{figure}[t]
	\begin{center}
		\includegraphics[width=0.95\linewidth, height=0.9\linewidth]{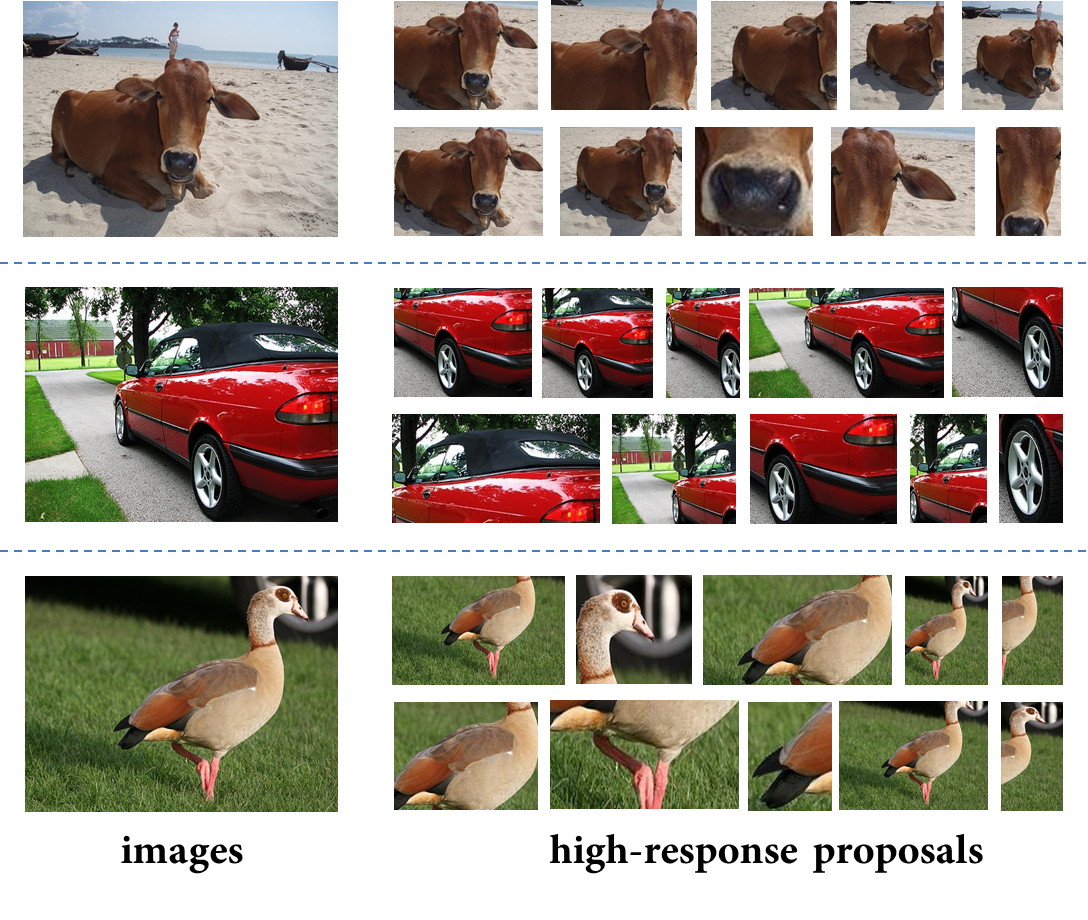}
	\end{center}
	\vspace{-0.7cm}
	\caption{ An illustration of candidate proposals with the highest responses to the corresponding class. Top $10$ proposals for each image are shown. The top-ranked proposals may contain context or only a key discriminative part of the object. However, these top-ranked proposals are mostly spatially concentrated around the true object instance. }
	\label{fig:top}
	\vspace{-0.3cm}
\end{figure}

\begin{figure}[t]
	\begin{center}
		\includegraphics[width=0.8\linewidth, height=0.8\linewidth]{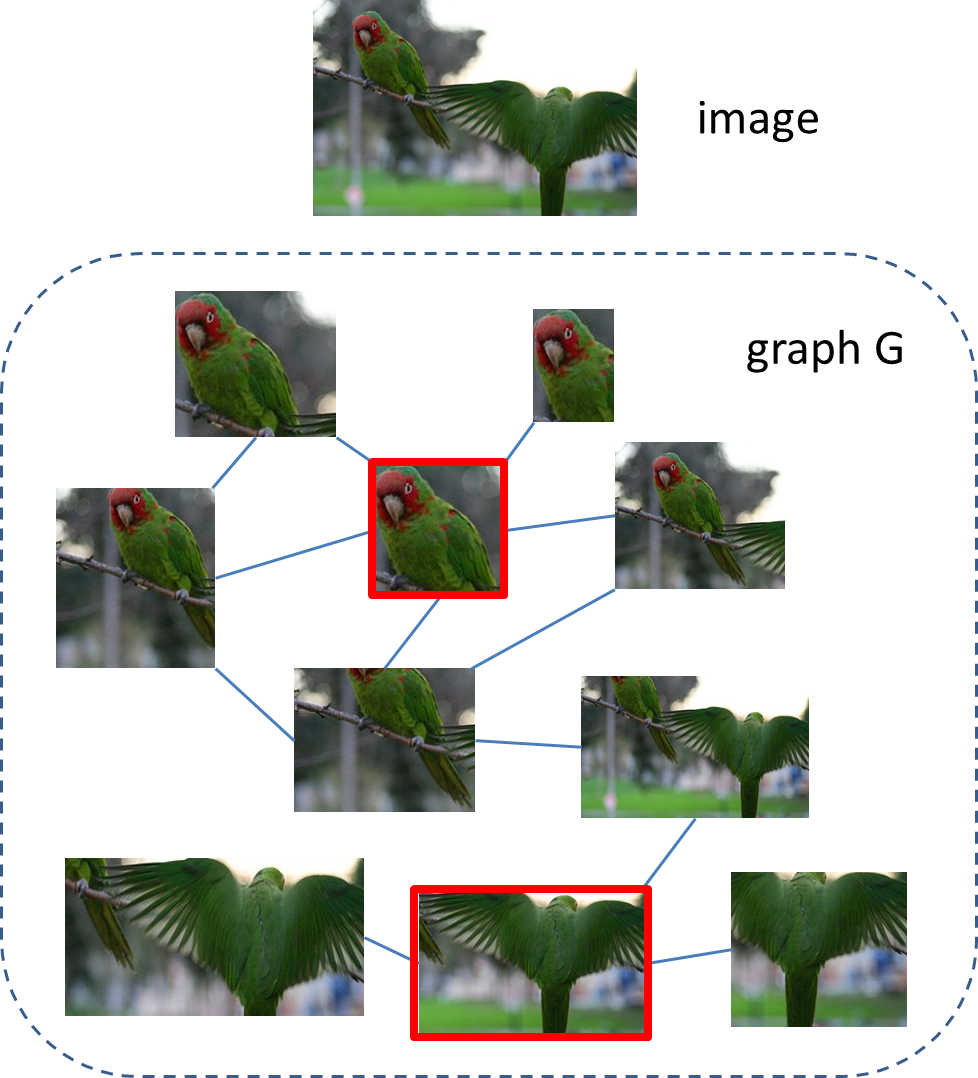}
	\end{center}
	\vspace{-0.5cm}
	\caption{ An illustration of graph $G$ whose nodes are the proposals in the $N$-candidate proposal pool. Each candidate proposal is connected to the others with IoU $\geq 0.5$ in this example. By dense subgraph discovery, two spatially concentrated proposals are selected among all the proposals, framed in red boxes.}
	\label{fig:graph}
	\vspace{-0.5cm}
\end{figure}
\vspace{-0.3cm}

\subsubsection{Reliable Seed Proposal Generation}

After image-to-object transferring, the top $N$ proposals with the highest predicted responses to the target class are selected as confident candidate proposals. However, high-response does not imply tight spatial coverage of the true object. Our experimental observation demonstrates that the proposals with some context or containing only the key discriminative part also have high responses to the target class in the above image-to-object transferring. Another key observation is that although some proposals contain part of the object or context, they may crowd the object (see Fig.~\ref{fig:top}). To incorporate the spatial correlation, we formulate it as a dense subgraph discovery (DSD) problem, \textit{i.e.}, selecting the most spatially concentrated ones in the candidate proposal pool that contains the $N$ high-response proposals.

Mathematically, let $G=(V,E)$ be an undirected unweighted graph whose nodes $V$ correspond to the top $N$ high-response proposals. The edges $E = \{e(v_i,v_j)\}$ are formed by connecting each proposal (node) to its neighboring proposals which have Intersection-over-Union (IoU) larger than a pre-defined threshold $T$. The visualization of an example graph $G$ is shown in Fig.~\ref{fig:graph}. We propose a greedy algorithm to discover the dense subgraph of $G$. The greedy algorithm iteratively selects the node with a greatest degree (number of connections to other nodes) and then prunes the node as well as all its connected neighbors. The algorithm repeats the finding-pruning iterations  until the number of the remained nodes is less than a pre-defined number  $k$. All the pruned nodes in the iterations form the dense subgraph. The procedure is detailed in Algorithm \ref{alg:1}.

\begin{algorithm}
	\caption{Dense Subgraph Discovery over Graph $G$} \label{alg:1}
	\begin{algorithmic}
		\STATE \textbf{Input:} An undirected graph $G = (V,E)$.
		\STATE \textbf{Initialization:} $V^{\prime} = \varnothing$.
		\WHILE{$|V|\textgreater k$}
		\STATE $v_{\max} = \arg\max_{i} d_i$, where  $d_i = \sum_{j \in V} e(v_i, v_j)$;
		\STATE $V_{\text{neighbor}} =  \{v | e(v,v_{\max}) =1\}$;
		\STATE $V^{\prime} = V^{\prime} \cup  \{v_{\max}\}$;
		\STATE $V = V\backslash V_{\text{neighbor}}$;
		\ENDWHILE
		\STATE \textbf{Output:}  A set of nodes $V^{\prime}$ constituting the dense subgraph.
	\end{algorithmic}
\end{algorithm}

Compared to other two ways of selecting spatially concentrated proposals, \emph{i.e.}, clustering and non-maximal suppression (NMS), DSD has the following appealing advantages. First, it can provide an adaptive number of proposals instead of requiring a pre-specified fixed number as clustering. This is highly desired in solving the WSL problem as images may have different numbers of object instances. Second, DSD does not rely on the predicted response, avoiding the unfavorable case, in which poor localized proposals with the highest responses are selected. This is a common issue with NMS, which cannot filter the proposals containing only a key discriminative part or context.

Among the selected spatially concentrated proposals, the one with the highest predicted response to the target class is selected as the seed positive sample for this image.

\subsection{Online Supportive Sample Harvesting}

After obtaining the seed positive proposals, we further seek higher-quality positive samples by taking advantage of the object-level CNN detector. In particular, we implement self-taught learning to improve the ability of the object-level regional CNN detector progressively.

We propose a novel online supportive sample harvesting (OSSH) strategy to progressively harvest the high-quality positive samples such that the quality of positive samples can be significantly improved. In this way, the ability of the detector can be substantially enhanced with the provided new informative samples. Fast R-CNN is used as our regional CNN detector. We observe that a regional CNN detector (Fast R-CNN) trained on seed samples is sufficiently powerful for selecting the most confident tight positives for further training itself.

Alternating between training and re-localization shares the similar spirit with the usual MIL that continuously updates SVM to mine high-quality positive samples. Although more powerful by using Fast R-CNN, one risk is that it is easily trapped in poor local optima caused by poor initial seeds due to its stronger fitting capacity.

\begin{figure}
	\captionsetup[subfigure]{labelformat=empty}	
	\centering
	\subfloat[]{\raisebox{0.26cm}{\includegraphics[width=0.15   \linewidth,height=0.34 \linewidth]{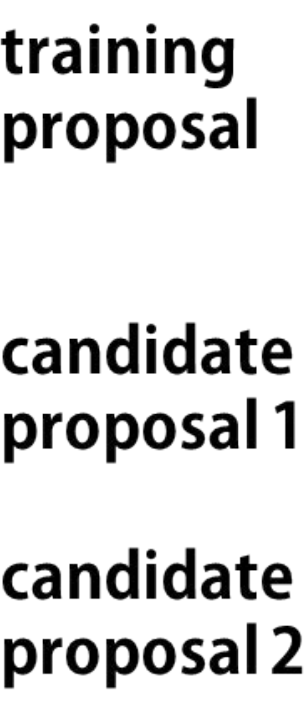}}
	}
	\hspace{0.01 \linewidth}
	\subfloat[]{\includegraphics[width=0.17   \linewidth,height=0.4  \linewidth]{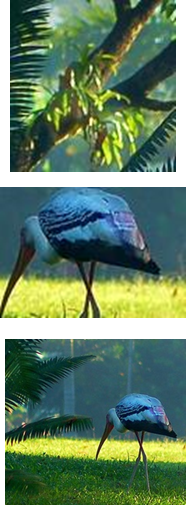}
	}
	\subfloat[]{\raisebox{-0.4cm}{\includegraphics[width=0.62   \linewidth,height=0.45  \linewidth]{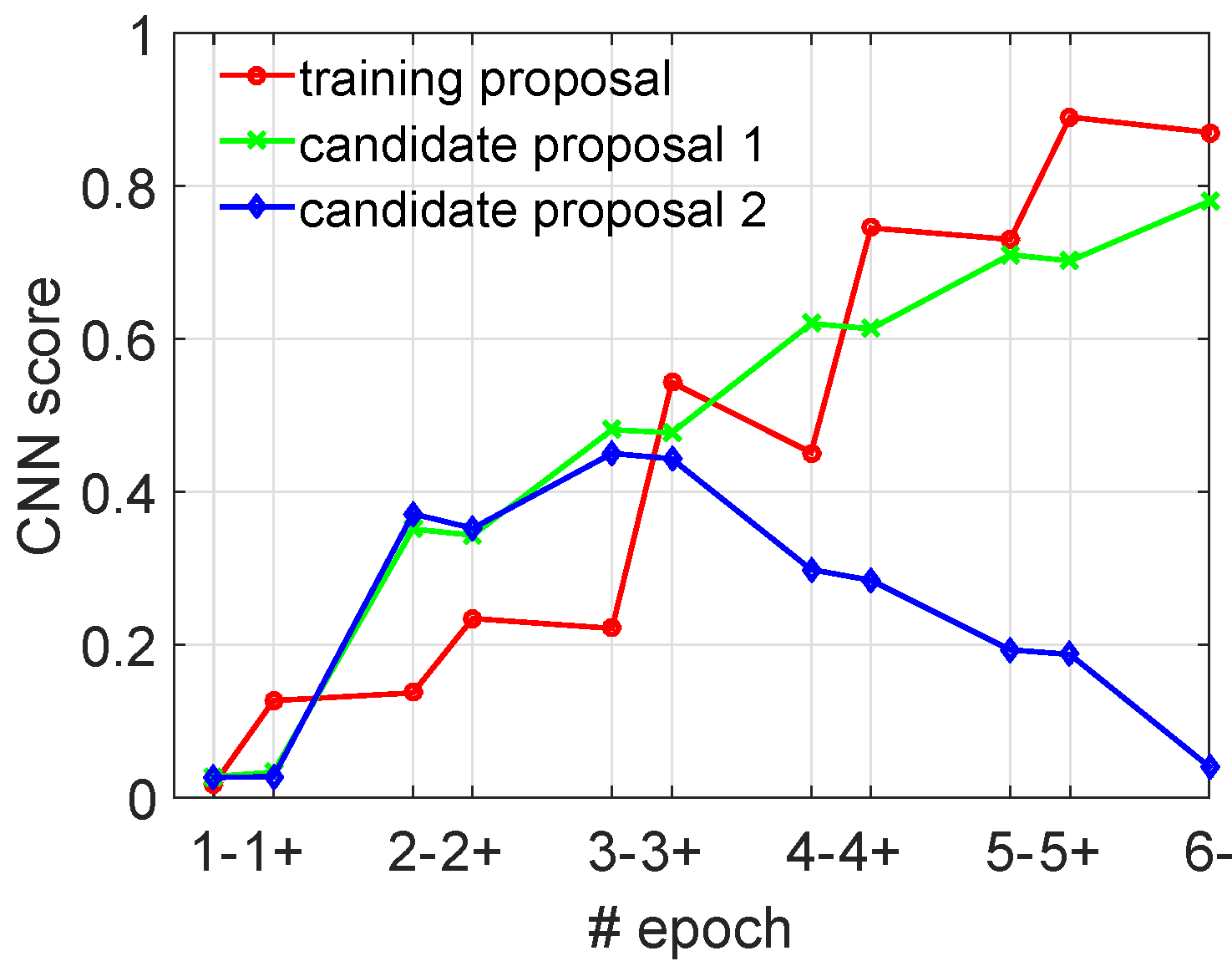}}
	}
	\\
	\vspace{-0.7cm}
	\subfloat[]{\includegraphics[width=0.176   \linewidth,height=0.035  \linewidth]{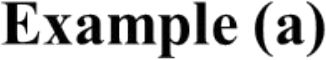}
	}
	\\
	\vspace{-0.5cm}
	\subfloat[]{\raisebox{0.18cm}{\includegraphics[width=0.15   \linewidth,height=0.33  \linewidth]{three_proposals_text2.pdf}}
	}
	\hspace{0.01 \linewidth}
	\subfloat[]{\includegraphics[width=0.17   \linewidth,height=0.4  \linewidth]{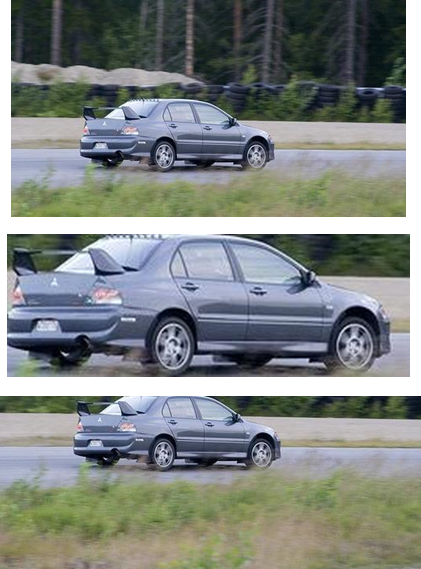}
	}
	\subfloat[]{\raisebox{-0.4cm}{\includegraphics[width=0.62   \linewidth,height=0.45  \linewidth]{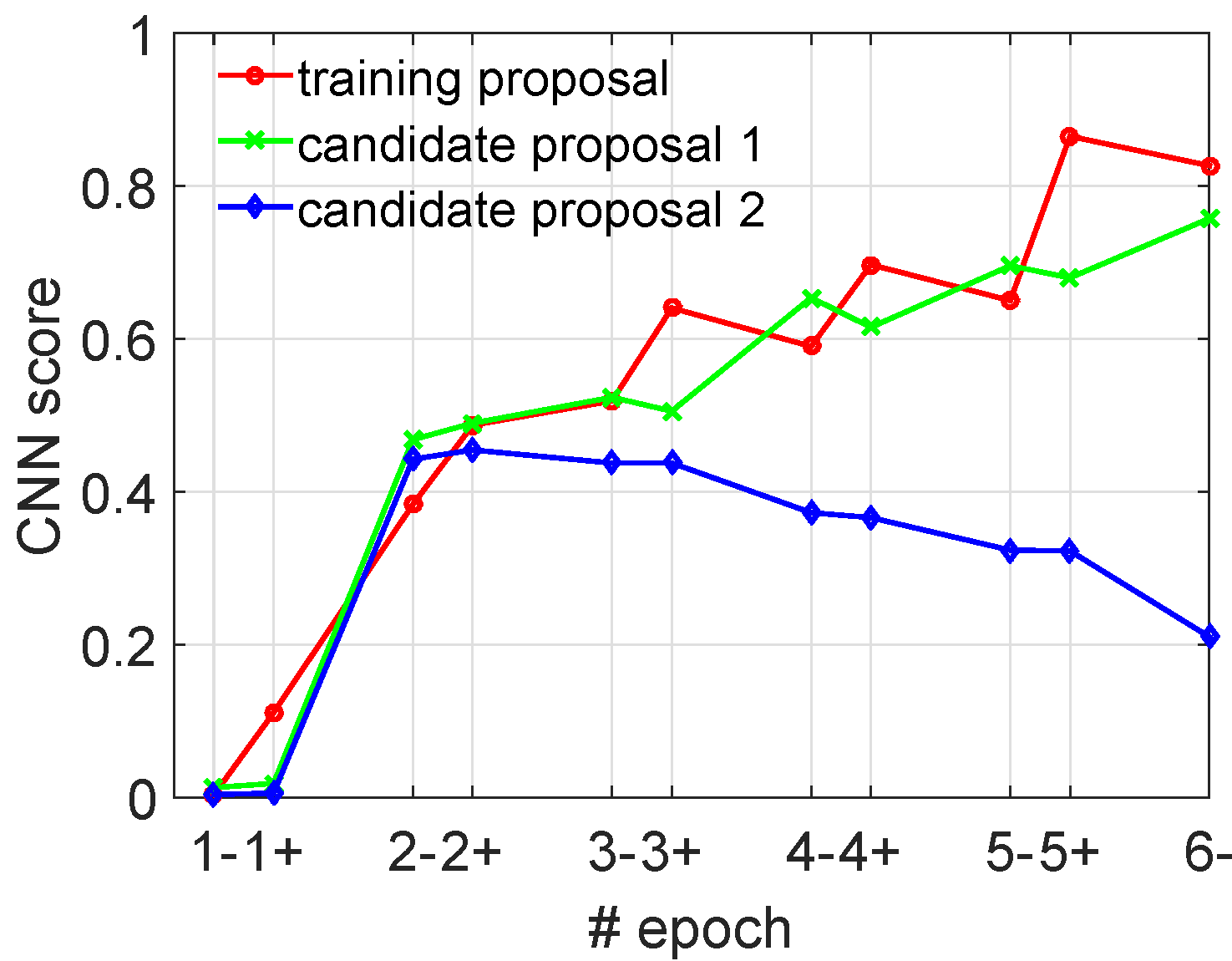}}
	}
	\\
	\vspace{-0.7cm}
	\subfloat[]{\includegraphics[width=0.176   \linewidth,height=0.035  \linewidth]{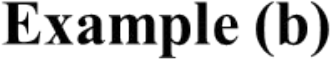}
	}
	\\
	\vspace{-0.5cm}
	\caption{ CNN score on the target class vs. number of epochs during training  Fast R-CNN for different proposals. The training proposals are the seed positive samples to train Fast R-CNN. ``$1$-'' and ``$1$+'' indicate the CNN score right before and after training on this image in the $1$st epoch, respectively. Similar meanings apply to the symbols in other epochs. High-quality proposals which are not used as training samples mainly gain score improvement from the increasing detection ability of Fast R-CNN, while the score improvement of false positive training samples mostly comes from the overfitting to themselves. }
	\label{fig:scores}
	\vspace{-0.4cm}
\end{figure}

To address this issue, we propose to online select the most confident and tight positive samples based on  \emph{relative  improvement} (RI) of output CNN scores, instead of relying on the static  absolute CNN score at certain training iterations. Specifically, for a training image, we rank all of its $N$ proposals in a descending order of RI over the last epoch. The proposal with the maximal RI is chosen as the positive training sample for the current epoch. For an image,  we denote the Fast R-CNN predicted score for the $i$-th proposal at the $t$-th epoch (after training Fast R-CNN on this image) as $A_{i}^{t}$. To compute the RI, we also denote  its Fast R-CNN score at the $(t{+}1)$-th epoch (but before training Fast R-CNN on this image) as $B_{i}^{t+1}$. Then among the $N$ candidate proposals, the proposal $P_{t+1}^{*}$ with the largest RI is selected for the $(t{+}1)$-th training epoch:
\begin{equation*}
	P_{t+1}^{*} = \arg\max_{i}(B_{i}^{t+1}-A_{i}^{t}).
\end{equation*}

We propose to use RI for proposal selection based on the following observations on the WSL problem. The high predicted score of a proposal may result from model overfitting to this proposal or the increasing detection ability of the Fast R-CNN model. We need to untangle these two factors as the former is not desired. Bad seed samples hardly obtain RI from the increasing detection ability of Fast R-CNN during training. In contrast, high-quality positive samples not selected as seeds mostly gain RI due to the improved detection ability of the model. Therefore, RI is a reliable metric for identifying high-quality positive samples.

Fig. \ref{fig:scores} shows intuitive examples to justify the  observations. In the Example (a) of Fig. \ref{fig:scores}, the score of the false initial training proposal gains improvement mostly from the overfitting to itself, and can hardly increase during training on other images (\emph{e.g.}, ``$1$+'' to ``$2$-'', ``$2$+'' to ``$3$-''), especially in later epochs (\emph{e.g.},  ``$3$+'' to ''$4$-'', ``$4$+'' to ``$5$-''). The high-quality candidate proposal (\emph{i.e.},  candidate proposal $1$) gains score improvement mostly during training on other images. The score of the low-quality candidate proposal (\emph{i.e.},  candidate proposal $2$ which contains context) improves during the increasing of the generalization power of the CNN model in early epochs (\emph{e.g.},  ``$1$+'' to ``$2$-''), but decreases in later epochs (\emph{e.g.},  ``$3$+'' to ``$4$-'', ``$4$+'' to ``$5$-'') when the CNN gains strong discrimination between the target class and background. In the Example (b) of Fig. \ref{fig:scores}, the low-quality seed training proposal has large score improvement when training on other images in early epochs (\emph{e.g.},  ``$1$+'' to ``$2$-''), similar to candidate proposal $2$, but can only gain score improvement from the overfitting to itself in later epochs.

Therefore, RI from the increasing detection ability of Fast R-CNN reliably reflects the quality of the proposal. To ensure the adequate positive samples from other images for training between two consecutive training on this image, \emph{e.g.}, at the $t$-th and $(t{+}1)$-th epoch, we fix the order of training images fed into the network in each epoch. This guarantees the model to be trained by all the rest images of the target class between two consecutive training on the particular image.

Finally, we introduce negative rejection (NR) performed after several epochs of online supportive sample harvesting (OSSH).  Specifically, we  perform NR by ranking all the positive samples with the highest predicted score from Fast R-CNN in each image in the order of their predicted CNN scores, and then remove $10\%$ samples with the minimal CNN scores and their corresponding images in the subsequent  Fast R-CNN training. This is inspired by the observation that even the best positive samples selected from the difficult positive images are of unsatisfactory quality (low IoU to true objects).

For data augmentation, apart from the selected proposals with the maximal relative score improvement, all the proposals in this image that overlap with the selected proposal by IoU $\geq 0.5$ are also treated as positives to train the detector at that epoch. The proposals which have IoU $\in [0.1, 0.5)$ overlap with the selected proposal are negative samples.

\begin{table*}[htbp]
	\scriptsize
	\newcommand{\tabincell}[2]{\begin{tabular}{@{}#1@{}}#2\end{tabular}}
	\setlength{\tabcolsep}{3.6pt}
	\renewcommand{\arraystretch}{1.3}
	\centering
	\caption{Correct localization (CorLoc) (\%) of our method and other state-of-the-art methods on the PASCAL  $2007$ \emph{trainval} set. OSSH1 performs OSSH only in the $2$nd epoch, OSSH2 performs OSSH in the $2$nd and $3$rd epochs, and OSSH3 performs OSSH in the $2$nd, $3$rd and $4$th epochs. }
	\begin{tabular}{l|cccccccccccccccccccc|c}
		\hline
		method & aero  & bike  & bird  & boat  & bottle & bus   & car   & cat   & chair & cow   & table & dog   & horse & mbike & person & plant & sheep & sofa  & train & tv    & Avg. \\
		\hline
		Cinbis \emph{et al.} \cite{cinbis2015multifold} & 57.2 & 62.2 & 50.9 & 37.9 & 23.9 & 64.8 & 74.4 & 24.8 & 29.7 & 64.1 & 40.8 & 37.3 & 55.6 & 68.1 & \J{25.5} & 38.5 & 65.2 & 35.8 & 56.6 & 33.5 & 47.3 \\
		Bilen \emph{et al.} \cite{bilen2015weakly} & 66.4 & 59.3 & 42.7 & 20.4 & 21.3 & 63.4 & 74.3 & 59.6 & 21.1 & 58.2 & 14.0 & 38.5 & 49.5 & 60.0 & 19.8 & 39.2 & 41.7 & 30.1 & 50.2 & 44.1 & 43.7\\
		Wang \emph{et al.} \cite{wang2014weakly} & 80.1 & 63.9 & 51.5 & 14.9 & 21.0 & 55.7 & 74.2 & 43.5 & 26.2 & 53.4 & 16.3 & 56.7 & 58.3 & 69.5 & 14.1 & 38.3 & 58.8 & 47.2 & 49.1 & 60.9 & 48.5 \\
		Kantorov \emph{et al.} \cite{kantorov2016contextlocnet} & \J{83.3} & 68.6 & 54.7 & 23.4 & 18.3 & 73.6 & 74.1 & 54.1 & 8.6 & 65.1 & \J{47.1} & 59.5 & 67.0 & 83.5 & 35.3 & 39.9 & 67.0 & \J{49.7} & 63.5 & 65.2 & 55.1 \\
		Li \emph{et al.} \cite{li2016domain} & 78.2 & 67.1 & \J{61.8} & 38.1 & \J{36.1} & 61.8 & 78.8 & 55.2 & \J{28.5} & \J{68.8} & 18.5 & 49.2 & \J{64.1} & 73.5 & 21.4 & \J{47.4} & 64.6 & 22.3 & 60.9 & 52.3 & 52.4 \\
		\hline
		HCP & 54.4  & 37.2  & 42.1  & 28.1  & 13.8  & 47.8  & 49.6  & 40.6  & 16.4  & 38.7  & 13.8  & 34.5  & 22.2  & 36.4  & 10.8  & 36.4  & 42.3  & 20.8  & 46.1  & 49.3 & 34.1 \\
		HCP+DSD & 56.9  & 36.0  & 45.4  & 26.5  & 15.7  & 49.8  & 54.5  & 53.1  & 15.9  & 45.6  & 13.4  & 37.5  & 38.1  & 42.1  & 16.2  & 34.2  & 45.4  & 29.7  & 55.6  & 46.1 & 37.9 \\
		HCP+DSD+OSSH1 & 70.2  & 60.0  & 53.9  & 26.1  & 28.3  & 58.9  & 75.4  & 58.9  & 14.8  & 63.4  & 17.9  & 52.6  & 51.7  & 67.0  & 19.7  & 46.3  & 63.9  & 42.4  & 67.0  & 65.1 & 50.2 \\
		HCP+DSD+OSSH2 & 73.9  & 56.0  & 52.1  & 26.9  & 34.0  & 66.6  & 80.0  & 59.5  & 13.1  & 70.2  & 22.9  & 55.7  & 60.6  & 83.8  & 22.0  & 51.5  & 71.1  & 50.4  & 71.2  & 74.4 & 54.9 \\
		HCP+DSD+OSSH3 & 72.7  & 55.3  & 53.0  & 27.8  & 35.2  & 68.6  & 81.9  & 60.7  & 11.6  & 71.6  & 29.7  & 54.3  & 64.3  & 88.2  & 22.2  & 53.7  & 72.2  & 52.6  & 68.9  & 75.5 & 56.1 \\
		\hline
	\end{tabular}%
	\label{tab:07_trainval_corloc}%
\end{table*}%

\begin{table*}[htbp]
	\scriptsize
	\newcommand{\tabincell}[2]{\begin{tabular}{@{}#1@{}}#2\end{tabular}}
	\setlength{\tabcolsep}{2.7pt}
	\renewcommand{\arraystretch}{1.3}
	\centering
	\caption{Detection average precision (AP) (\%) of  our method  and other state-of-the-art methods (trained on the PASCAL $2007$ \emph{trainval} set) on the PASCAL  $2007$ \emph{test} set. OSSH1, OSSH2 and OSSH3 have the same meanings as Table \ref{tab:07_trainval_corloc}. 07+12 means training on the PASCAL $2007$ \emph{trainval} and $2012$ \emph{trainval} sets.}
	\begin{tabular}{l|cccccccccccccccccccc|c}
		\hline
		method & aero  & bike  & bird  & boat  & bottle & bus   & car   & cat   & chair & cow   & table & dog   & horse & mbike & person & plant & sheep & sofa  & train & tv    & mAP \\
		\hline
		Cinbis \emph{et al.} \cite{cinbis2015multifold} & 38.1 & 47.6 & 28.2 & 13.9 & 13.2& 45.2 & 48.0 & 19.3 & 17.1 & 27.7 & 17.3 & 19.0 & 30.1 & 45.4 & \J{13.5} & 17.0 & 28.8 & 24.8 & 38.2 & 15.0 & 27.4 \\
		Song \emph{et al.} \cite{song2014learning} & 27.6 & 41.9 & 19.7 & 9.1 & 10.4 & 35.8 & 39.1 & 33.6 & 0.6 & 20.9 & 10.0 & 27.7 & 29.4 & 39.2 & 9.1 & 19.3 & 20.5 & 17.1 & 35.6 & 7.1 & 22.7 \\
		Bilen \emph{et al.} \cite{bilen2015weakly} & 46.2 & 46.9 & 24.1 & 16.4 & 12.2 & 42.2 & 47.1 & 35.2 & 7.8 & 28.3 & 12.7 & 21.5 & 30.1 & 42.4 & 7.8 & 20.0 & 26.8 & 20.8 & 35.8 & 29.6 & 27.7\\
		Wang \emph{et al.} \cite{wang2014weakly} & 48.9 & 42.3 & 26.1 & 11.3 & 11.9 & 41.3 & 40.9 & 34.7 & 10.8 & 34.7 & 18.8 & 34.4 & 35.4 & 52.7 & 19.1 & 17.4 & 35.9 & 33.3 & 34.8 & 46.5 & 31.6 \\
		Kantorov \emph{et al.} \cite{kantorov2016contextlocnet} & \J{57.1} & 52.0 & 31.5 & 7.6 & 11.5 & 55.0 & 53.1 & 34.1 & 1.7 & 33.1 & \J{49.2} & 42.0 & 47.3 & 56.6 & 15.3 & 12.8 & 24.8 & \J{48.9} & 44.4 & 47.8 & 36.3 \\
		Li \emph{et al.} \cite{li2016domain} & 54.5 & 47.4 & \J{41.3} & 20.8 & \J{17.7} & 51.9 & 63.5 & 46.1 & \J{21.8} & \J{57.1} & 22.1 & 34.4 & \J{50.5} & 61.8 & 16.2 & \J{29.9} & 40.7 & 15.9 & 55.3 & 40.2 & 39.5 \\
		\hline
		HCP & 42.6  & 40.8  & 26.5  & 21.0    & 5.7   & 41.7  & 47.8  & 34.2  & 10.8  & 27.2  & 12.3  & 28.9  & 12.5  & 27.9  & 1.8   & 18.2  & 29.0    & 12.5  & 45.5  & 47.1  & 26.7 \\
		HCP+DSD & 45.7  & 41.0    & 26.8  & 23.1  & 5.0     & 51.4  & 51.5  & 43.3  & 10.4  & 37.6  & 10.2  & 29.2  & 23.0    & 39.1  & 3.1   & 16.8  & 33.5  & 13.6  & 47.2  & 40.5  & 29.6 \\
		HCP+DSD+OSSH1 & 52.5  & \J{56.9}  & 35.5  & 18.5  & 13.8  & 59.5  & 62.4  & 51.7  & 7.0     & 53.1  & 14.9  & 38.3  & 34.6  & 60.0    & 5.7   & 15.1  & 49.7  & 36.0    & 55.7  & 54.6  & 38.8 \\
		HCP+DSD+OSSH2 & 52.9  & 53.6  & 32.4  & 20.3  & 14.8  & 59.2  & 64.8  & 50.3  & 3.3   & 51.2  & 16.7  & 42.5  & 44.4  & 62.9  & 6.1   & 19.1  & 47.2  & 42.0    & \J{57.1}  & 62.4  & 40.2 \\
		HCP+DSD+OSSH3 & 49.6  & 47.0    & 33.6  & 21.7  & 15.7  & 60.4  & 66.0    & 51.7  & 5.6   & 54.1  & 24.5  & 38.4  & 45.2  & 65.0    & 6.1   & 18.5  & \J{53.3}  & 46.0    & 52.5  & 61.5  & 40.8 \\
		HCP+DSD+OSSH3+NR & 52.2  & 47.1  & 35.0    & \J{26.7}  & 15.4  & \J{61.3}  & \J{66.0}    & \J{54.3}  & 3.0     & 53.6  & 24.7  & \J{43.6}  & 48.4  & \J{65.8}  & 6.6   & 18.8  & 51.9  & 43.6  & 53.6  & \J{62.4}  & \J{41.7} \\
		HCP+DSD+OSSH3+NR (07+12) & 54.2  & 52.0    & 35.2  & 25.9  & 15.0    & 59.6  & 67.9  & 58.7  & 10.1  & 67.4  & 27.3  & 37.8  & 54.8  & 67.3  & 5.1   & 19.7  & 52.6  & 43.5  & 56.9  & 62.5  & 43.7 \\
		\hline
	\end{tabular}%
	\label{tab:07_test}%
\end{table*}%

\begin{table*}[htbp]
	\scriptsize
	\newcommand{\tabincell}[2]{\begin{tabular}{@{}#1@{}}#2\end{tabular}}
	\setlength{\tabcolsep}{3.25pt}
	\renewcommand{\arraystretch}{1.3}
	\centering
	\caption{Detection average precision (AP) (\%) of our method  and other state-of-the-art methods (trained on the PASCAL $2012$ \emph{train} set) on the PASCAL  $2012$ \emph{val} set. OSSH1, OSSH2 and OSSH3 have the same meanings as Table \ref{tab:07_trainval_corloc}. }
	\begin{tabular}{l|cccccccccccccccccccc|c}
		\hline
		method & aero  & bike  & bird  & boat  & bottle & bus   & car   & cat   & chair & cow   & table & dog   & horse & mbike & person & plant & sheep & sofa  & train & tv    & mAP \\
		\hline
		Li \emph{et al.} \cite{li2016domain} & -- & -- & -- & -- & -- & -- & -- & -- & -- & -- & -- & -- & -- & -- & -- & -- & -- & -- & -- & -- & 29.1 \\
		\hline
		HCP & 49.3  & 33.3  & 24.7  & 14.0    & 11.8  & 37.9  & 30.2  & 35.7  & 6.9   & 26.6  & 6.9   & 25.4  & 14.1  & 29.4  & 1.1   & 18.1  & 25.7  & 13.4  & 44.1  & 45.4  & 24.7 \\
		HCP+DSD & 55.3  & 39.3  & 25.3  & 14.3  & 10.6  & 50.4  & 35.6  & 45.4  & 11.4  & 31.3  & 2.3   & 30.6  & 29.7  & 35.3  & 5.0     & 14.2  & 28.1  & 13.8  & 47.1  & 41.1  & 28.3 \\
		HCP+DSD+OSSH1 & 60.7  & 54.0    & 36.5  & 14.4  & 19.5  & 57.5  & 45.5  & 47.7  & 11.1  & 39.9  & 2.8   & 43.4  & 38.2  & 55.5  & 4.3   & 18.6  & 40.5  & 31.1  & 56.6  & 52.0    & 36.5 \\
		HCP+DSD+OSSH2 & 57.7  & 55.9  & 34.8  & 17.4  & 18.3  & 57.8  & 48.6  & 51.0    & 9.7   & 40.8  & 7.2   & 42.5  & 47.2  & 62.2  & 4.6   & 18.4  & 43.0    & 36.8  & 55.7  & 57.8  & 38.4 \\
		HCP+DSD+OSSH3 & 61.0    & 53.8  & 30.3  & 18.1  & 18.6  & 57.4  & 51.1  & 53.1  & 6.1   & 40.7  & 12.1  & 38.2  & 48.2  & 65.5  & 4.8   & 20.9  & 45.5  & 34.0    & 54.1  & 57.3  & 38.5 \\
		HCP+DSD+OSSH3+NR & 60.9  & 53.3  & 31.0    & 16.4  & 18.2  & 58.2  & 50.5  & 55.6  & 9.1   & 42.1  & 12.1  & 43.4  & 45.3  & 64.6  & 7.4   & 19.3  & 44.8  & 39.3  & 51.4  & 57.2  & 39.0 \\
		\hline
	\end{tabular}%
	\label{tab:12_val}%
\end{table*}%

\begin{table*}[htbp]
	\scriptsize
	\newcommand{\tabincell}[2]{\begin{tabular}{@{}#1@{}}#2\end{tabular}}
	\setlength{\tabcolsep}{3.6pt}
	\renewcommand{\arraystretch}{1.3}
	\centering
	\caption{Correct localization (CorLoc) (\%) of our method and other state-of-the-art ones on the PASCAL $2012$ \emph{trainval} set.}
	\begin{tabular}{l|cccccccccccccccccccc|c}
		\hline
		method & aero  & bike  & bird  & boat  & bottle & bus   & car   & cat   & chair & cow   & table & dog   & horse & mbike & person & plant & sheep & sofa  & train & tv    & Avg. \\
		\hline
		Kantorov \emph{et al.} \cite{kantorov2016contextlocnet} & 78.3 & 70.8 & 52.5 & 34.7 & 36.6 & 80.0 & 58.7 & 38.6 & 27.7 & 71.2 & 32.3 & 48.7 & 76.2 & 77.4 & 16.0 & 48.4 & 69.9 & 47.5 & 66.9 & 62.9 & 54.8 \\
		\hline
		HCP+DSD+OSSH3 & 82.4  & 68.1  & 54.5    & 38.9  & 35.9  & 84.7  & 73.1  & 64.8  & 17.1   & 78.3  & 22.5  & 57.0  & 70.8  & 86.6  & 18.7   & 49.7  & 80.7  & 45.3  & 70.1  & 77.3  & 58.8 \\
		\hline
	\end{tabular}%
	\label{tab:12_trainval_corloc}%
\end{table*}%

\begin{table*}[htbp]
	\scriptsize
	\newcommand{\tabincell}[2]{\begin{tabular}{@{}#1@{}}#2\end{tabular}}
	\setlength{\tabcolsep}{2.7pt}
	\renewcommand{\arraystretch}{1.3}
	\centering
	\caption{Detection average precision (AP) (\%) of  our method  and other state-of-the-art methods (trained on the PASCAL $2012$ \emph{trainval} set) on the PASCAL $2012$ \emph{test} set. 07+12 means training on the PASCAL $2007$ \emph{trainval} and $2012$ \emph{trainval} sets.}
	\begin{tabular}{l|cccccccccccccccccccc|c}
		\hline
		method & aero  & bike  & bird  & boat  & bottle & bus   & car   & cat   & chair & cow   & table & dog   & horse & mbike & person & plant & sheep & sofa  & train & tv    & mAP \\
		\hline
		Kantorov \emph{et al.} \cite{kantorov2016contextlocnet} & 64.0 & 54.9 & 36.4 & 8.1 & 12.6 & 53.1 & 40.5 & 28.4 & 6.6 & 35.3 & 34.4 & 49.1 & 42.6 & 62.4 & 19.8 & 15.2 & 27.0 & 33.1 & 33.0 & 50.0 & 35.3 \\
		\hline
		HCP+DSD+OSSH3+NR & 60.8  & 54.2  & 34.1    & 14.9  & 13.1  & 54.3  & 53.4  & 58.6  & 3.7   & 53.1  & 8.3  & 43.4  & 49.8  & 69.2  & 4.1   & 17.5  & 43.8  & 25.6  & 55.0  & 50.1  & 38.3 \\
		HCP+DSD+OSSH3+NR (07+12) & 62.4  & 55.3  & 34.1    & 17.1  & 17.3  & 56.4  & 54.9  & 57.6  & 3.9   & 54.6  & 6.7  & 44.3  & 52.0  & 71.2  & 4.0   & 17.3  & 42.9  & 28.4  & 54.1  & 52.5  & 39.4 \\
		\hline
	\end{tabular}%
	\label{tab:12_test}%
\end{table*}%

\section{Experiments}
\subsection{Datasets and Evaluation Metrics}

We evaluate our approach on PASCAL VOC $2007$ and $2012$ datasets \cite{everingham2010pascal} which are the most widely-used benchmarks in weakly supervised object detection. For PASCAL $2007$, we train the model on the \emph{trainval} set (containing $5,011$ images) and evaluate on the \emph{test} set (containing $4,952$ images). For PASCAL $2012$, we first train the model on the \emph{train} set (containing $5,717$ images) and evaluate on the \emph{val} set (containing $5,823$ images). Additionally, we also train our model on the PASCAL $2012$ \emph{trainval} set (containing $11,540$ images) and evaluate on the \emph{test} set (containing $1,0991$ images).

We use two metrics in the evaluation of our approach. First, standard detection mean average precision (mAP) defined by \cite{everingham2010pascal} is evaluated on the PASCAL $2007$ \emph{test} set, PASCAL $2012$ \emph{val} set and PASCAL $2012$ \emph{test} set with their respective training models stated above.  Second, on the training sets (\emph{i.e.}, the PASCAL $2007$ \emph{trainval} set and PASCAL $2012$ \emph{trainval} set), we report Correct Localization (CorLoc) \cite{deselaers2012weakly} which is a standard metric for measuring localization accuracy on a training set. CorLoc is the percentage of images, where the most confident detected bounding box overlaps (IoU$\geq 0.5$) with a ground-truth box.

\subsection{Implementation Details}

We train the HCP multi-label classification model with the settings following \cite{wei2015hcp}. 
In all the experiments, $100$ proposals with the highest responses to the target class are chosen to form the candidate proposal pool to balance the performance and efficiency. In dense subgraph discovery, we fix the values of $T$ and $k$ to $0.8$ and $5$ for all the experiments, as it is empirically shown that the localization performance will not change much when $T$ is greater than $0.7$ or when $k$ ranges from $3$ to $8$. In the Fast R-CNN training with online supportive sample harvesting, the model is fine-tuned from the pre-trained model on ImageNet~\cite{deng2009imagenet}. The batch size is set to $2$ such that the overfitting to a certain image resulting from the training on that mini-batch is obvious. The order of training images is fixed in all the epochs. The learning rate is set to $0.001$ initially and decreased by a factor of $10$ after every $6$ epochs. We use the object proposals generated by Edge Boxes \cite{zitnick2014edge}, and adopt the VGG-16 network \cite{simonyan2014very} in the Fast R-CNN. 

\subsection{Ablation Studies}

To validate the effectiveness of our two components, \emph{i.e.}, dense subgraph discovery and online supportive sample harvesting, we conduct ablation studies by accumulatively adding each of them to our baseline, \emph{i.e.}, HCP. The baseline HCP selects the proposal with the highest response to the target class as the positive sample in each image.  In all the ablation versions of our method, Fast R-CNN is trained with the proposals with IoU$\geq0.5$ to their respective positive samples. From Table~\ref{tab:07_trainval_corloc}, one can observe that DSD improves CorLoc by nearly $4\%$ compared to  only using HCP to select positive proposals. OSSH1, OSSH2 and OSSH3 indicate performing online supportive sample harvesting in the first $1$, $2$ and $3$ epochs from the $2$nd epoch of training Fast R-CNN (note in the $1$st epoch, seed positives from DSD are used in training).~$12\%$ of improvement on CorLoc brought by OSSH1 shows that performing OSSH only $1$ time for a certain image  adequately discovers the tight positive proposal in the candidate pool. It can be seen that later OSSH has a less benefit to CorLoc than the OSSH in the $2$nd epoch,  showing that high-quality positive proposals gain consistent CNN score improvements in each of these epochs and thus can be easily picked out in the first time of OSSH. Table~\ref{tab:07_test} shows that mAP has similar trends to CorLoc. DSD and OSSH1 bring around $3\%$ and $9\%$ improvements in mAP respectively, validating their effectiveness. NR is also beneficial to the detector and contributes $1\%$ mAP improvement by discarding the false positives from the difficult images. Table \ref{tab:12_val} also shows significant improvements of mAP after adding DSD and OSSH to the baseline method on the PASCAL 2012 \emph{val} set.

To validate the advantage of using relative CNN score improvement, we conduct comparison experiments with using absolute CNN scores to harvest confident positive samples in OSSH. After epochs of OSSH, the proposals with the highest predicted score in each image are selected as confident positive samples. From Table \ref{tab:ablation}, it is found that relative score improvement consistently outperforms absolute CNN scores in all cases, especially when OSSH is performed in more epochs. Using absolute CNN scores, the improvements of OSSH in the later two epochs are much less than using relative score improvement.  This further demonstrates that the detector is more easily trapped in poor local optima when selecting positive samples based on absolute CNN scores, since the detector highly overfits seed positive samples and thus seed positive samples can obtain high predicted scores after the first $2$ epochs.

\begin{figure*}[t]
	\newcommand{\tabincell}[2]{\begin{tabular}{@{}#1@{}}#2\end{tabular}}
	\captionsetup[subfigure]{labelformat=empty}	
	\centering
	
	\subfloat[]{\includegraphics[width=0.17   \linewidth,height=0.11  \linewidth]{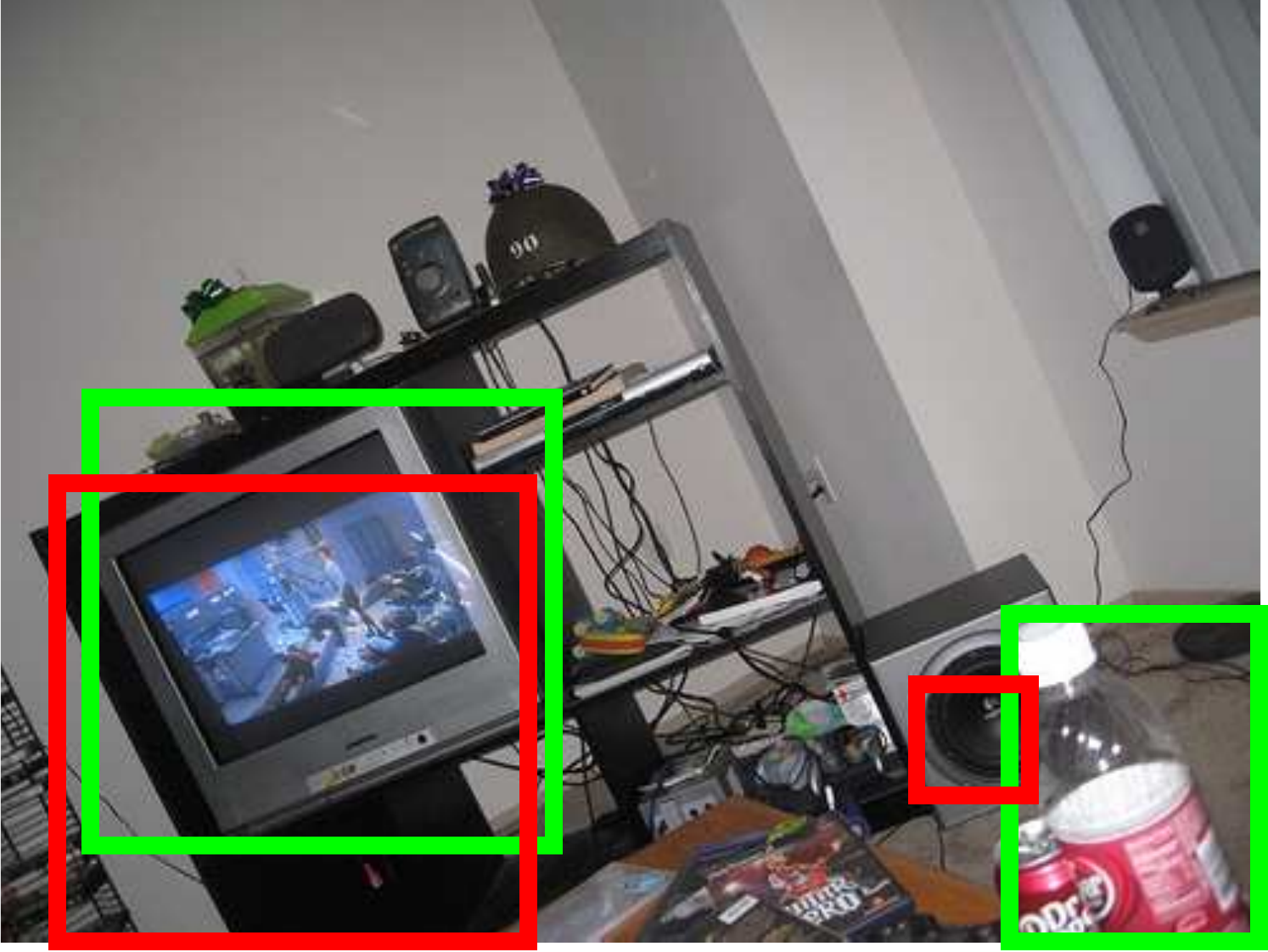}
	}
	\hspace{0.005 \linewidth}
	\subfloat[]{\includegraphics[width=0.17   \linewidth,height=0.11  \linewidth]{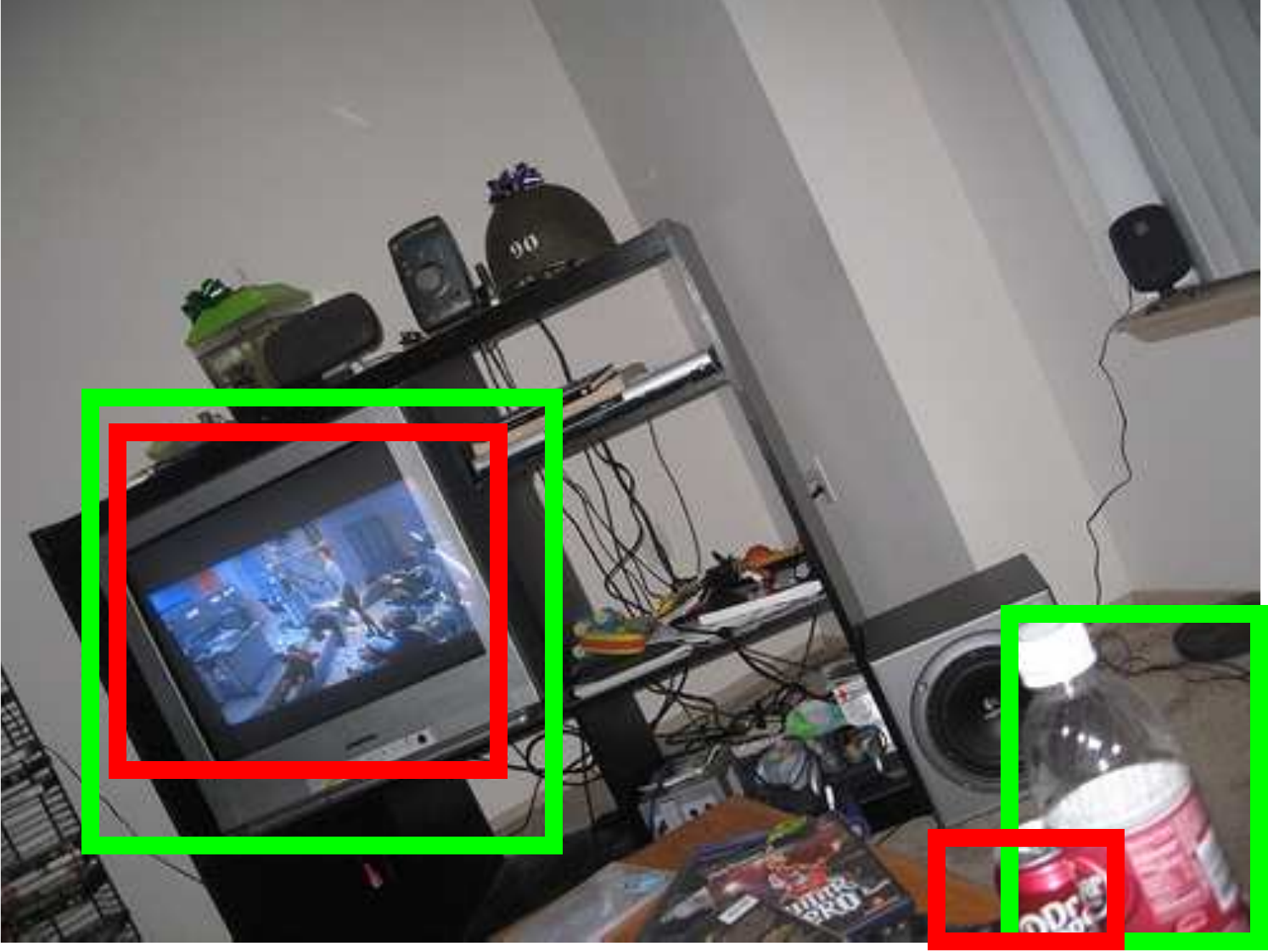}
	}
	\hspace{0.005 \linewidth}
	\subfloat[]{\includegraphics[width=0.17   \linewidth,height=0.11  \linewidth]{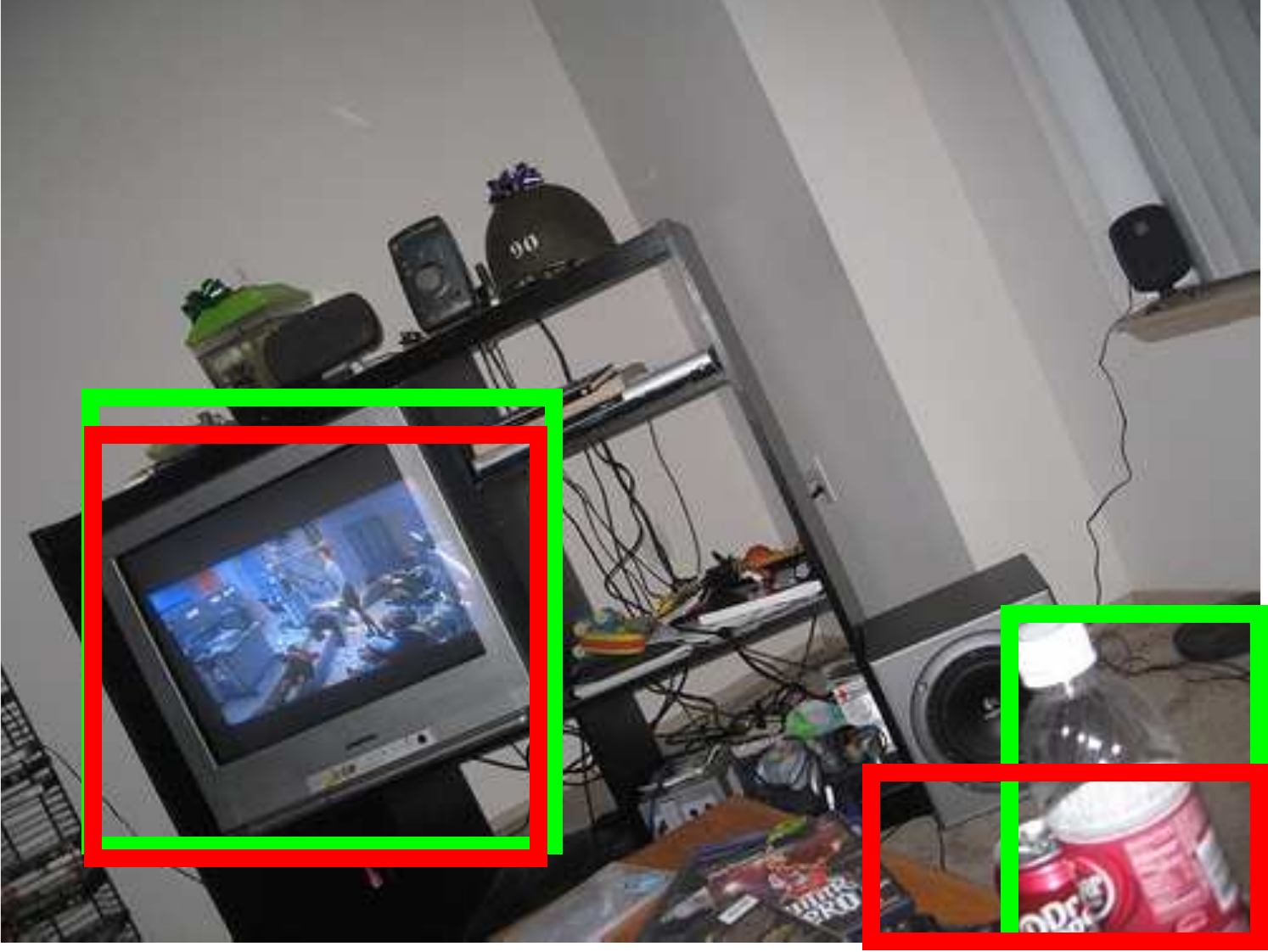}
	} 
	\hspace{0.005 \linewidth}
	\subfloat[]{\includegraphics[width=0.17   \linewidth,height=0.11  \linewidth]{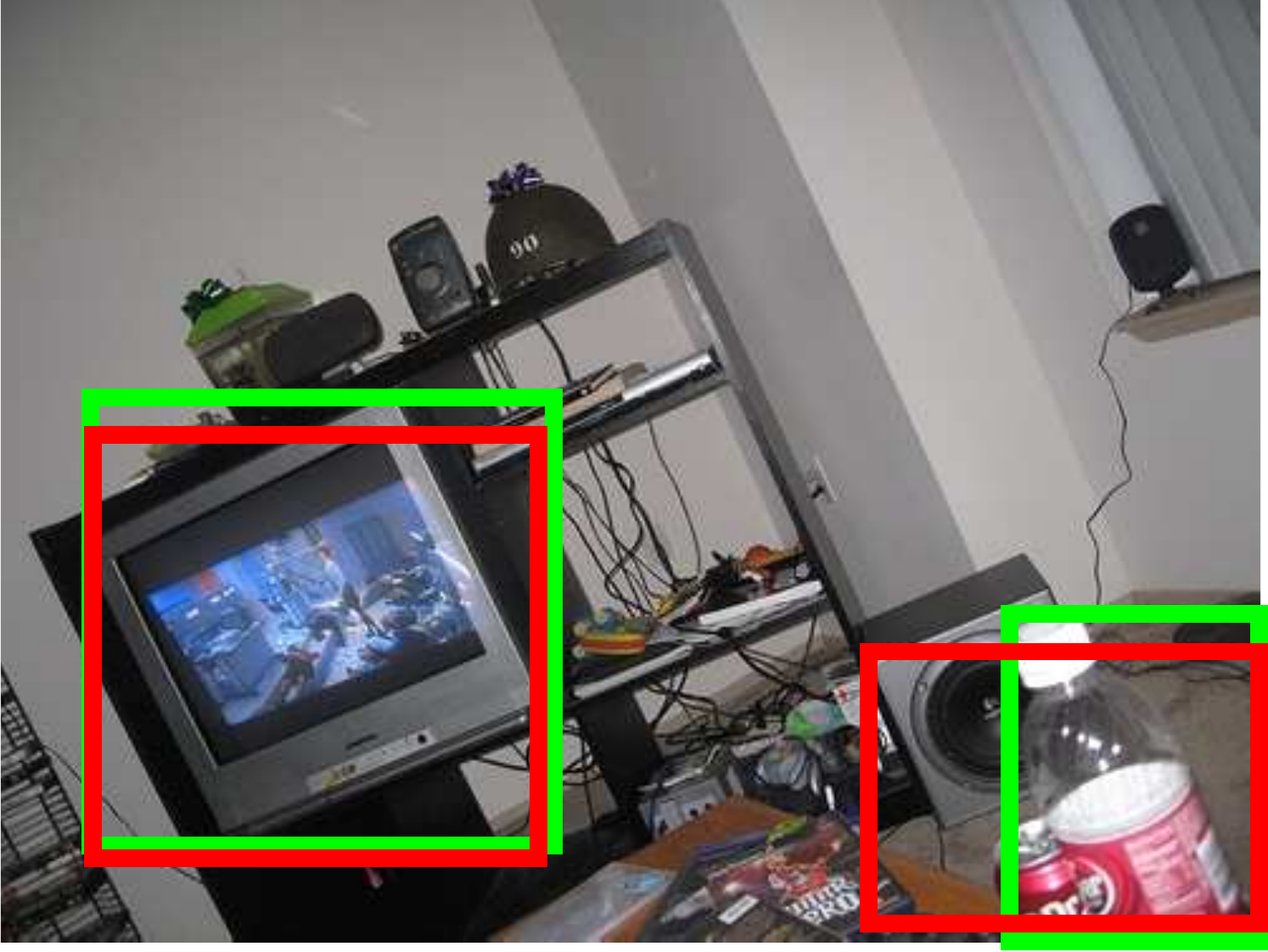}
	}
	\hspace{0.005 \linewidth}
	\subfloat[]{\includegraphics[width=0.17   \linewidth,height=0.11  \linewidth]{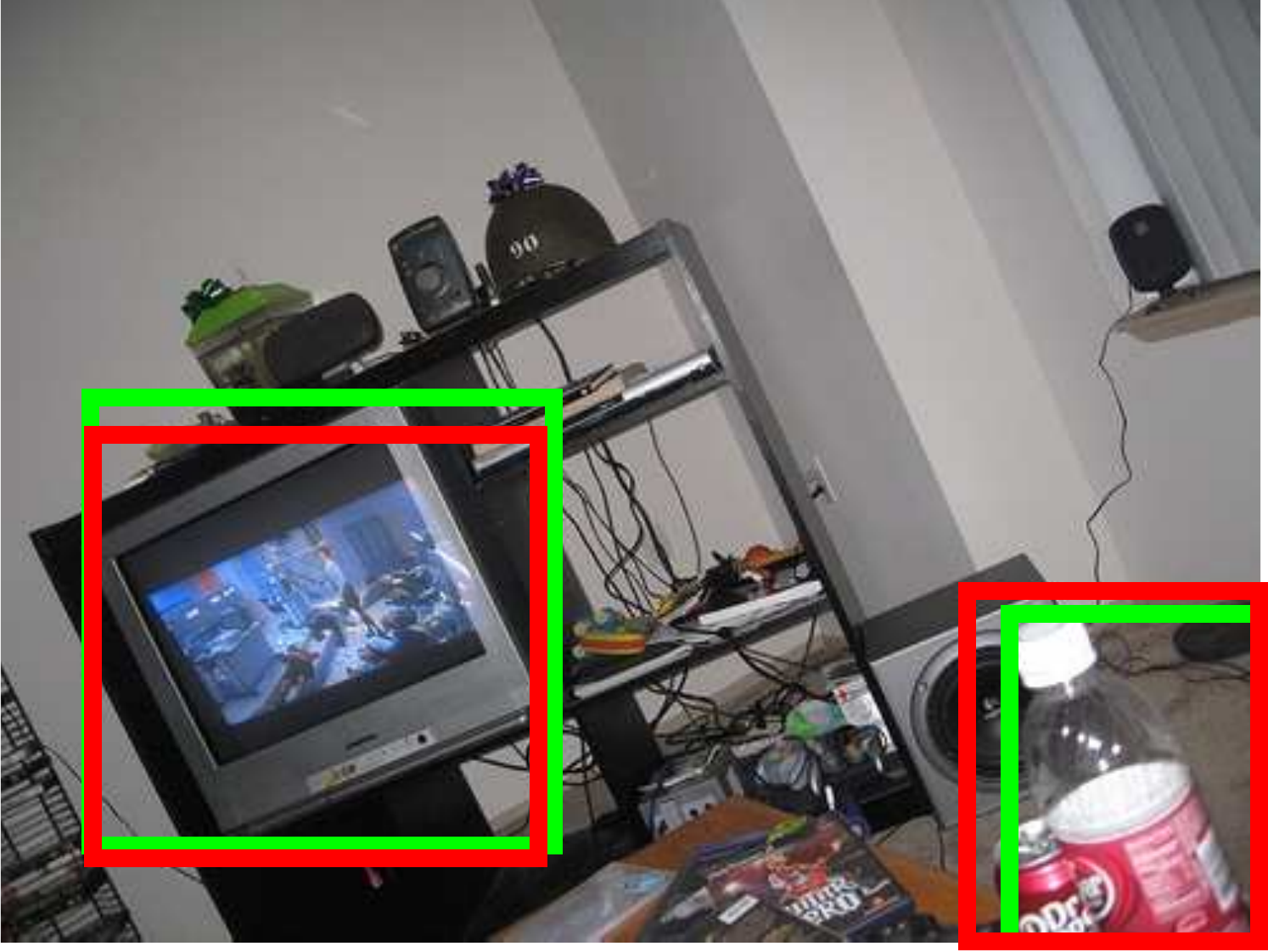}
	}
	\\
	
	\vspace{-0.6cm}
	\subfloat[]{\includegraphics[width=0.17   \linewidth,height=0.12  \linewidth]{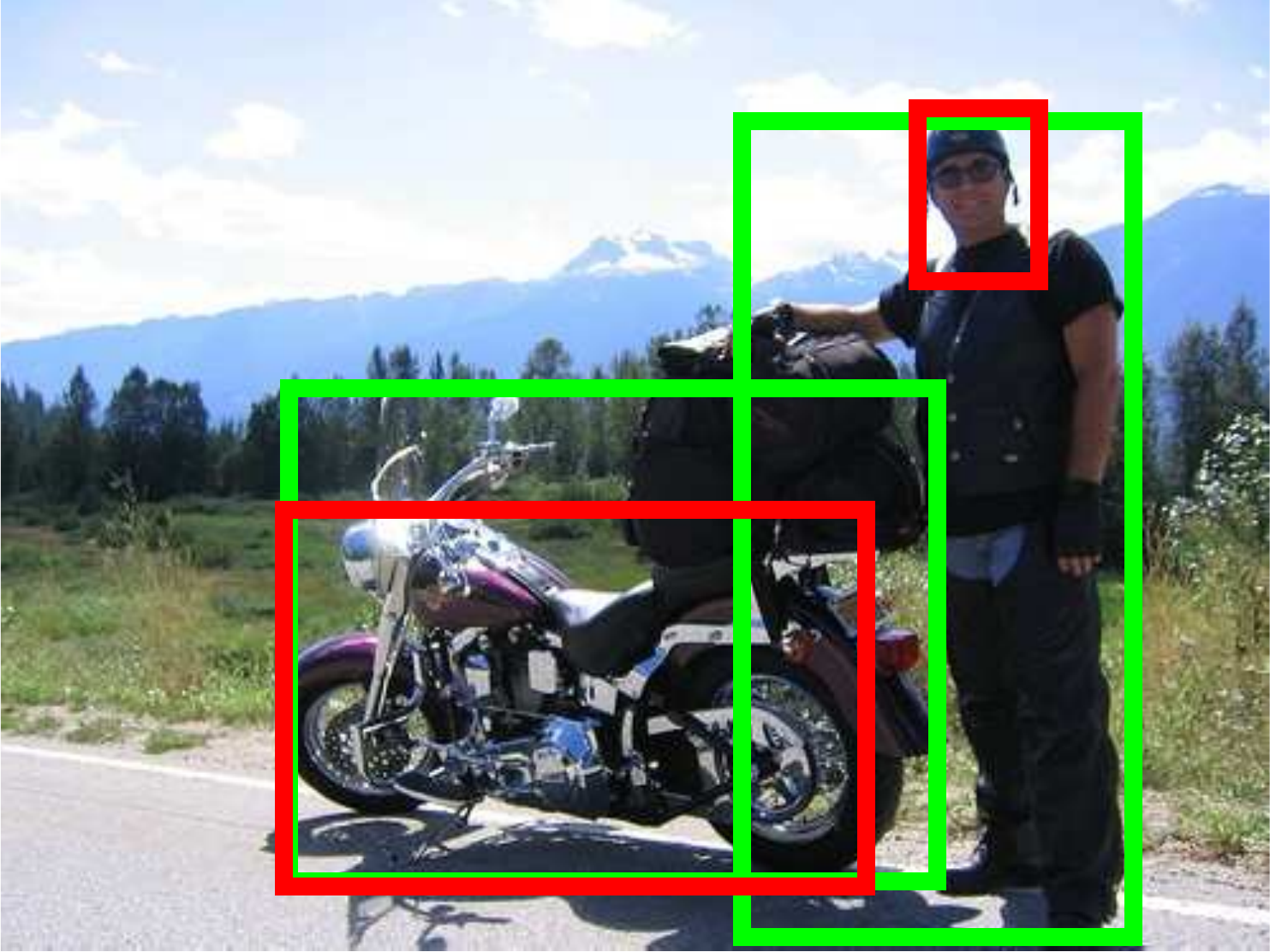}
	}
	\hspace{0.005 \linewidth}
	\subfloat[]{\includegraphics[width=0.17   \linewidth,height=0.12  \linewidth]{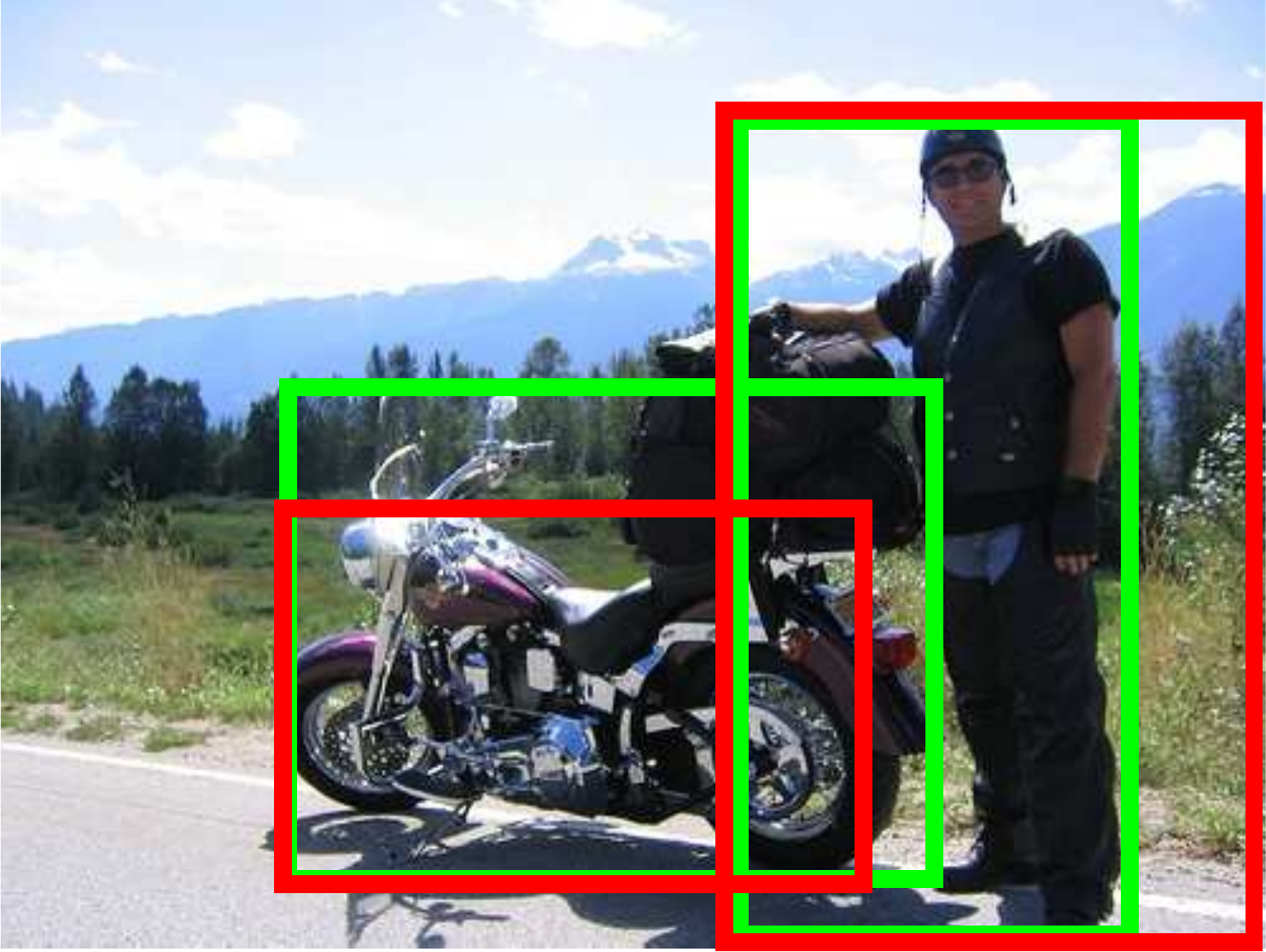}
	}
	\hspace{0.005 \linewidth}
	\subfloat[]{\includegraphics[width=0.17   \linewidth,height=0.12  \linewidth]{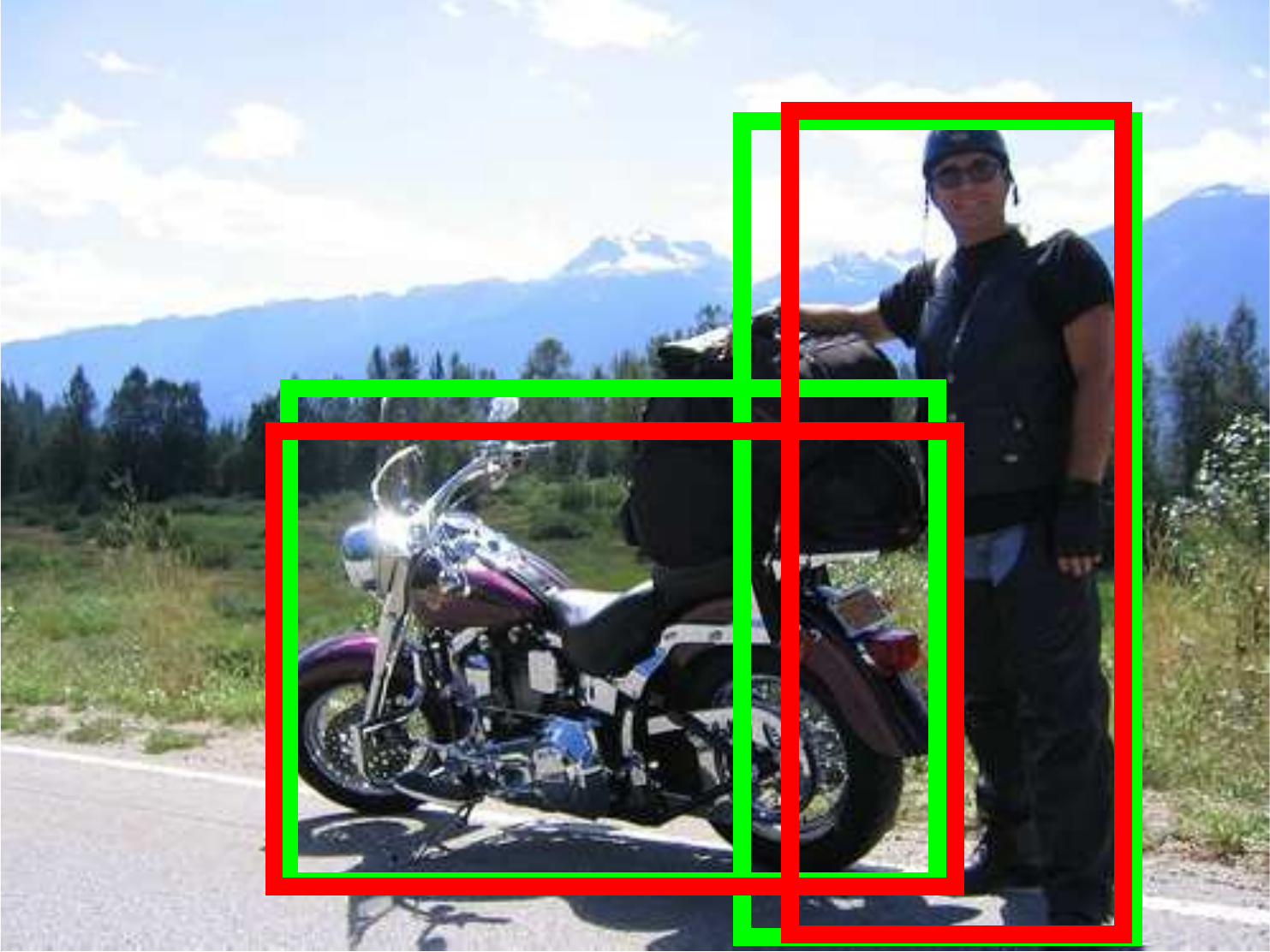}
	} 
	\hspace{0.005 \linewidth}
	\subfloat[]{\includegraphics[width=0.17   \linewidth,height=0.12  \linewidth]{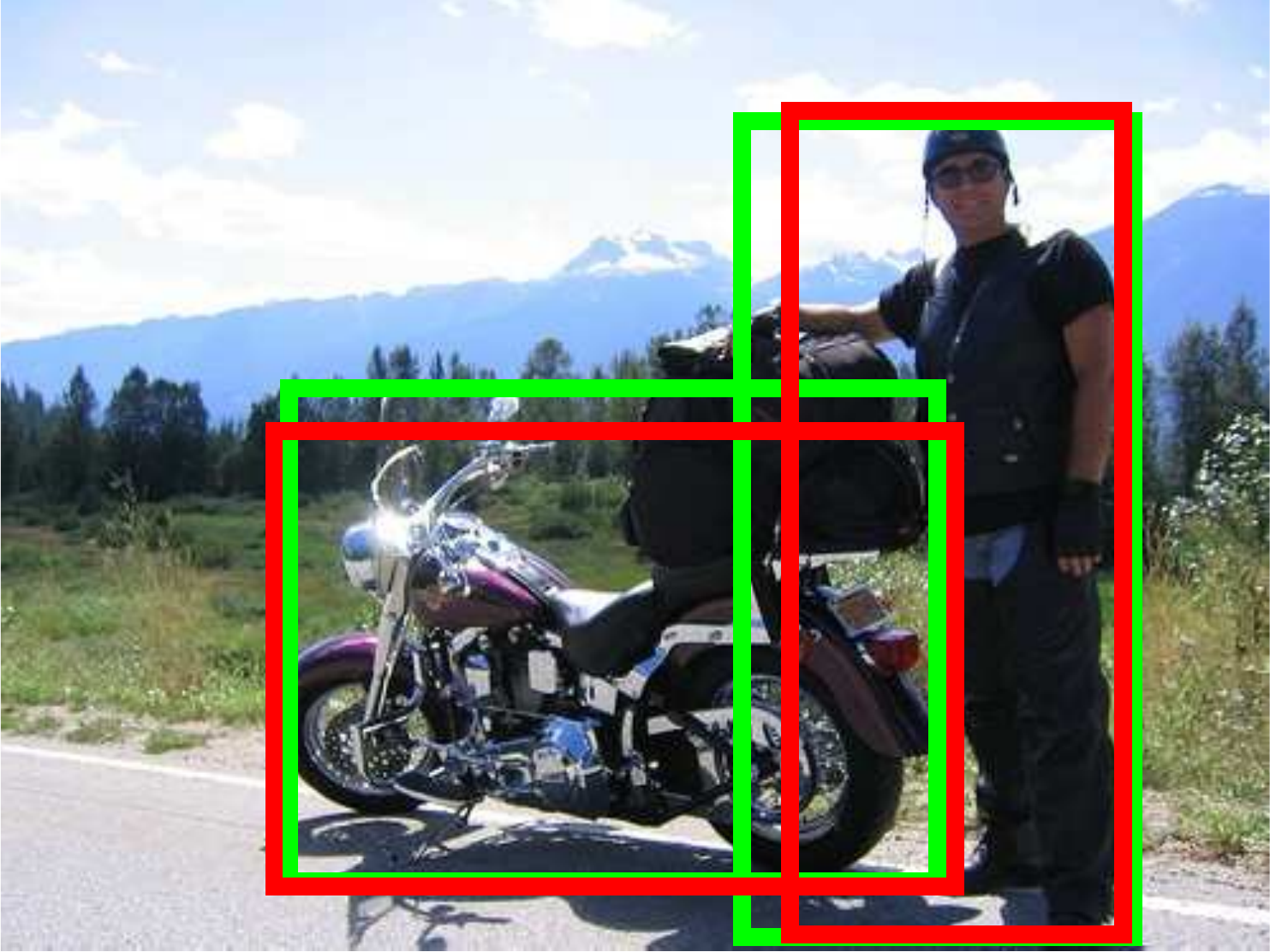}
	}
	\hspace{0.005 \linewidth}
	\subfloat[]{\includegraphics[width=0.17   \linewidth,height=0.12  \linewidth]{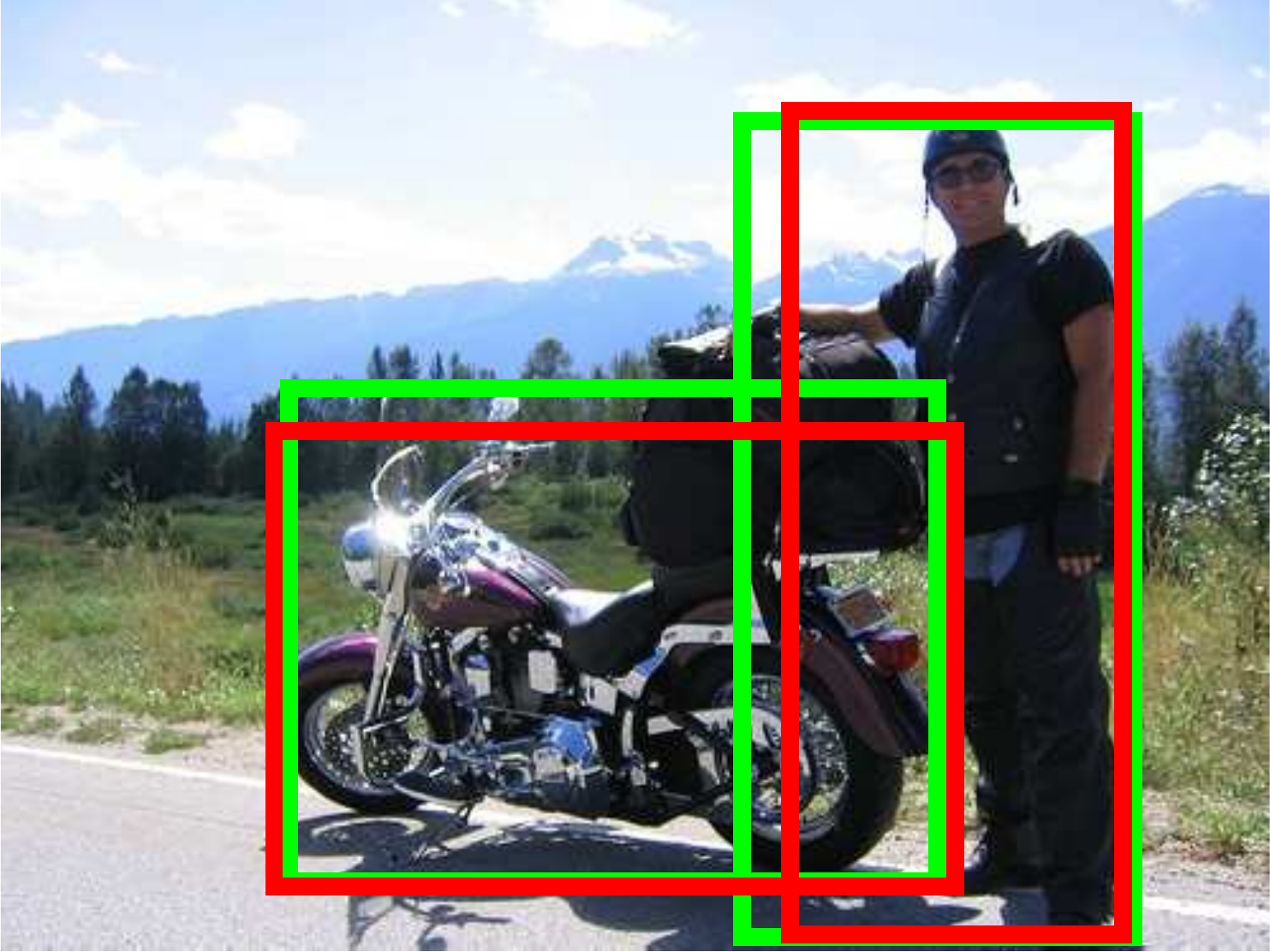}
	}
	\\
	\vspace{-0.6cm}
	\subfloat[]{\includegraphics[width=0.17   \linewidth,height=0.12  \linewidth]{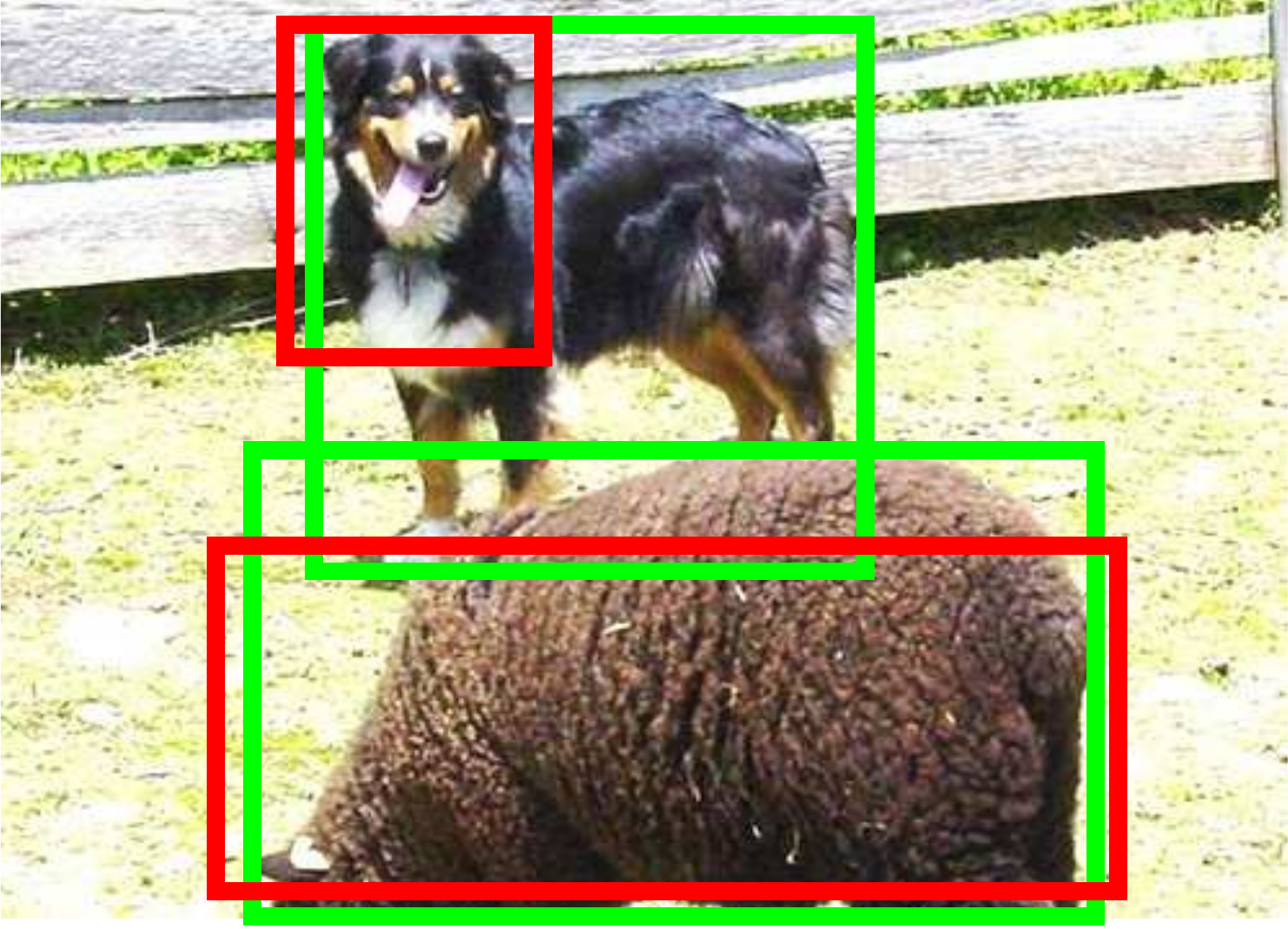}
	}
	\hspace{0.005 \linewidth}
	\subfloat[]{\includegraphics[width=0.17   \linewidth,height=0.12  \linewidth]{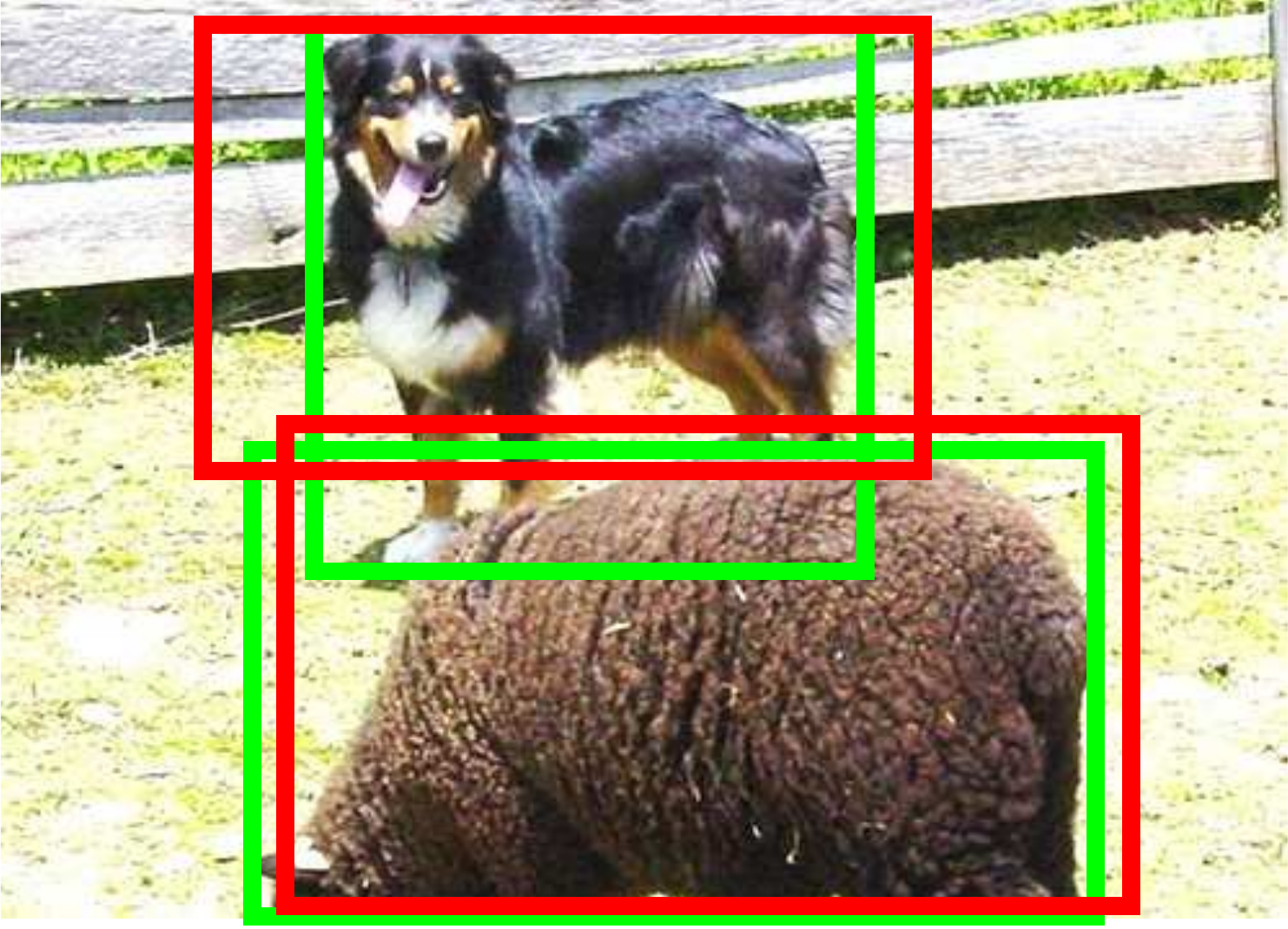}
	}
	\hspace{0.005 \linewidth}
	\subfloat[]{\includegraphics[width=0.17   \linewidth,height=0.12  \linewidth]{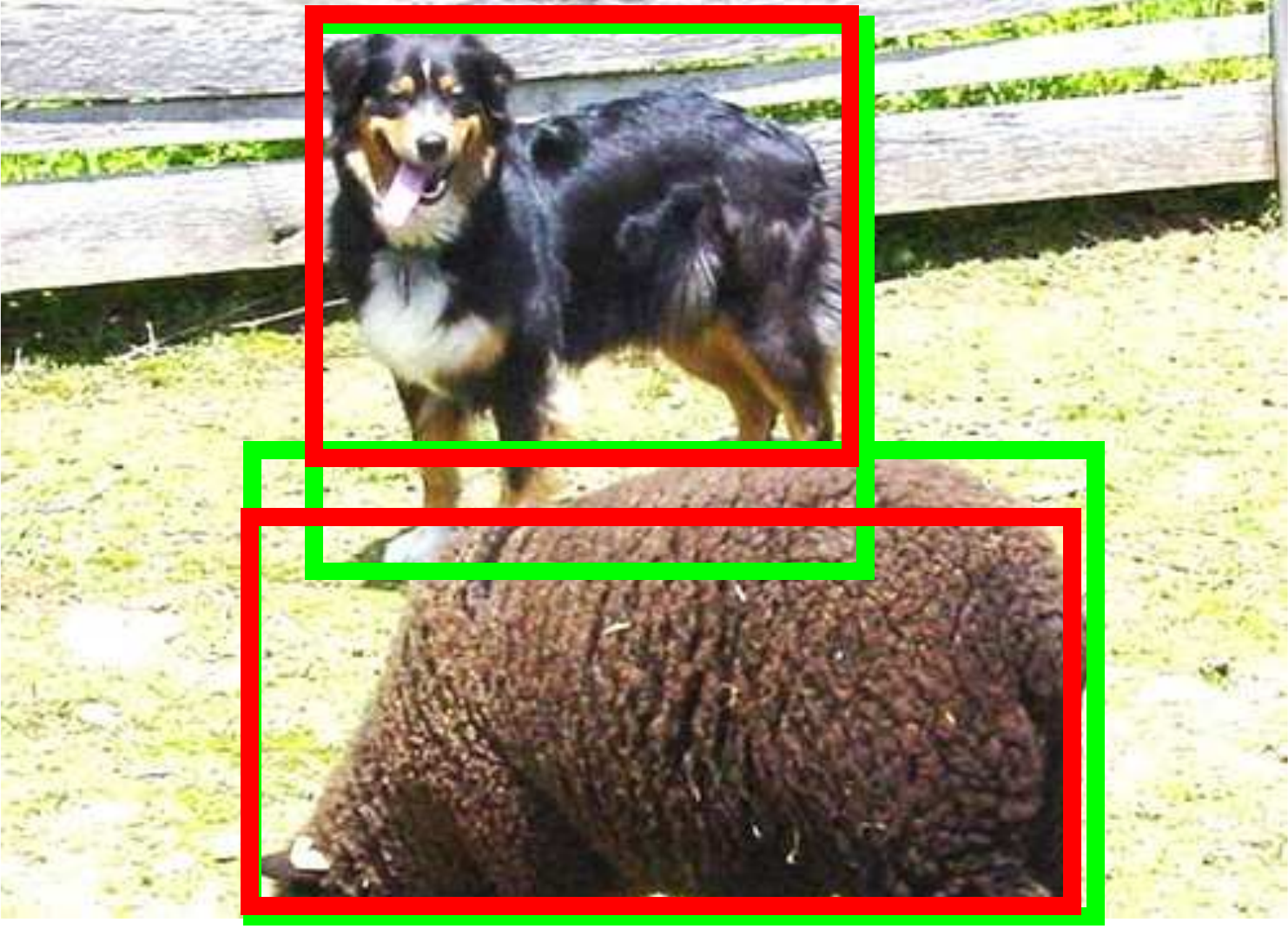}
	} 
	\hspace{0.005 \linewidth}
	\subfloat[]{\includegraphics[width=0.17   \linewidth,height=0.12  \linewidth]{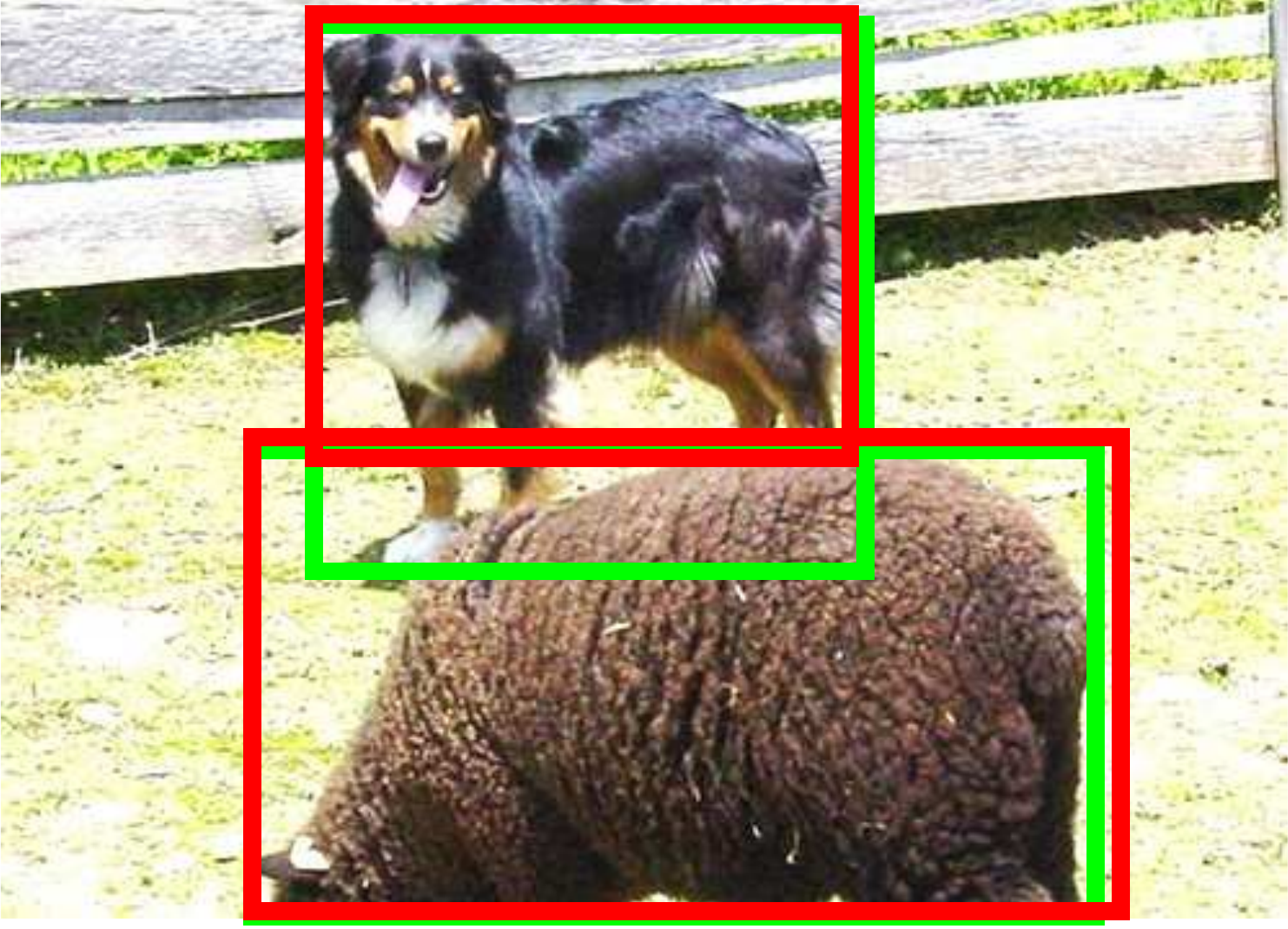}
	}
	\hspace{0.005 \linewidth}
	\subfloat[]{\includegraphics[width=0.17   \linewidth,height=0.12  \linewidth]{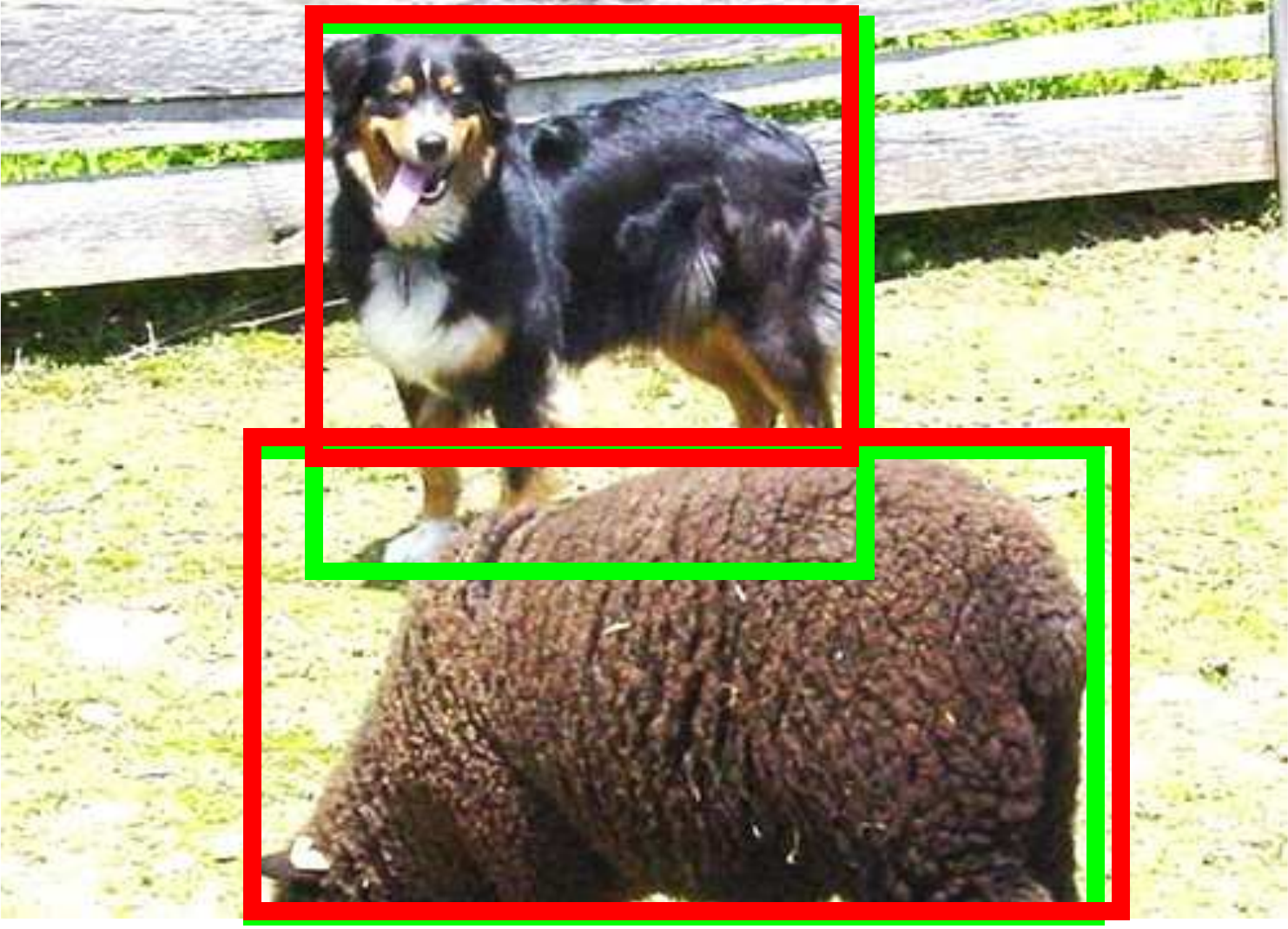}
	}
	\\
	\vspace{-0.6cm}
	\subfloat[HCP]{\includegraphics[width=0.14   \linewidth,height=0.14  \linewidth]{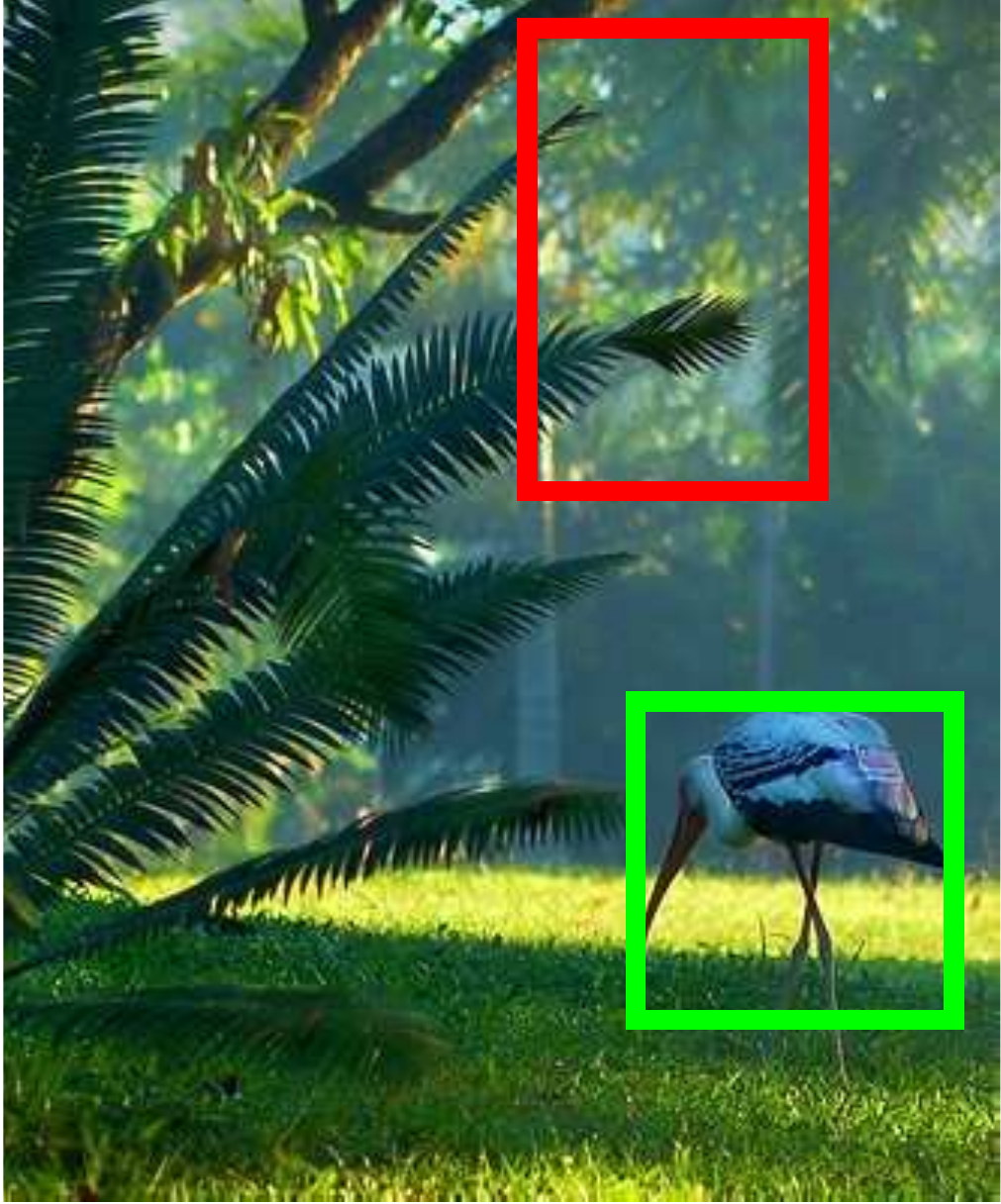}
	}
	\hspace{0.035 \linewidth}
	\subfloat[HCP+DSD]{\includegraphics[width=0.14   \linewidth,height=0.14  \linewidth]{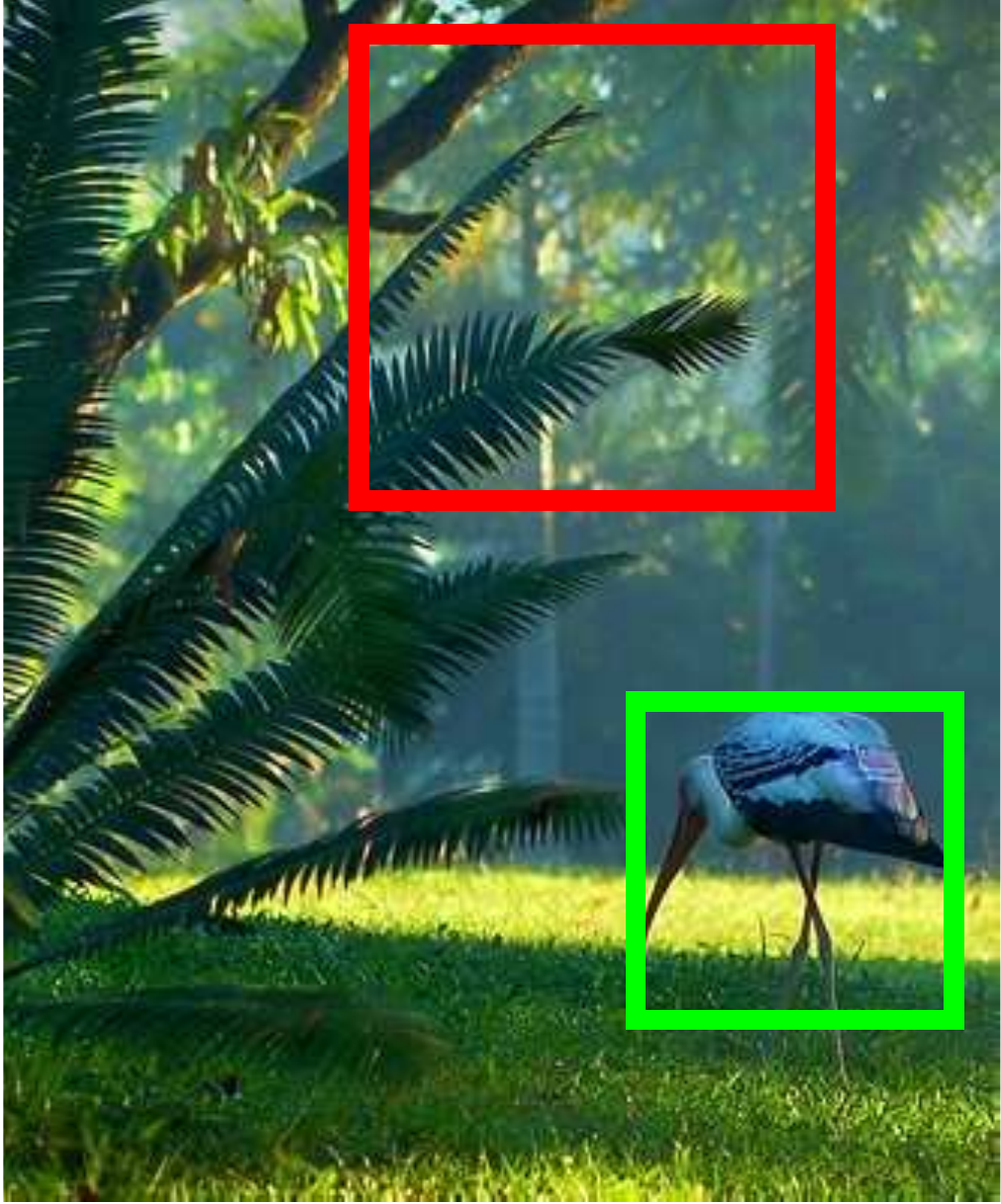}
	}
	\hspace{0.035 \linewidth}
	\subfloat[HCP+DSD, +OSSH1][HCP+DSD \\ +OSSH1]{\includegraphics[width=0.14   \linewidth,height=0.14  \linewidth]{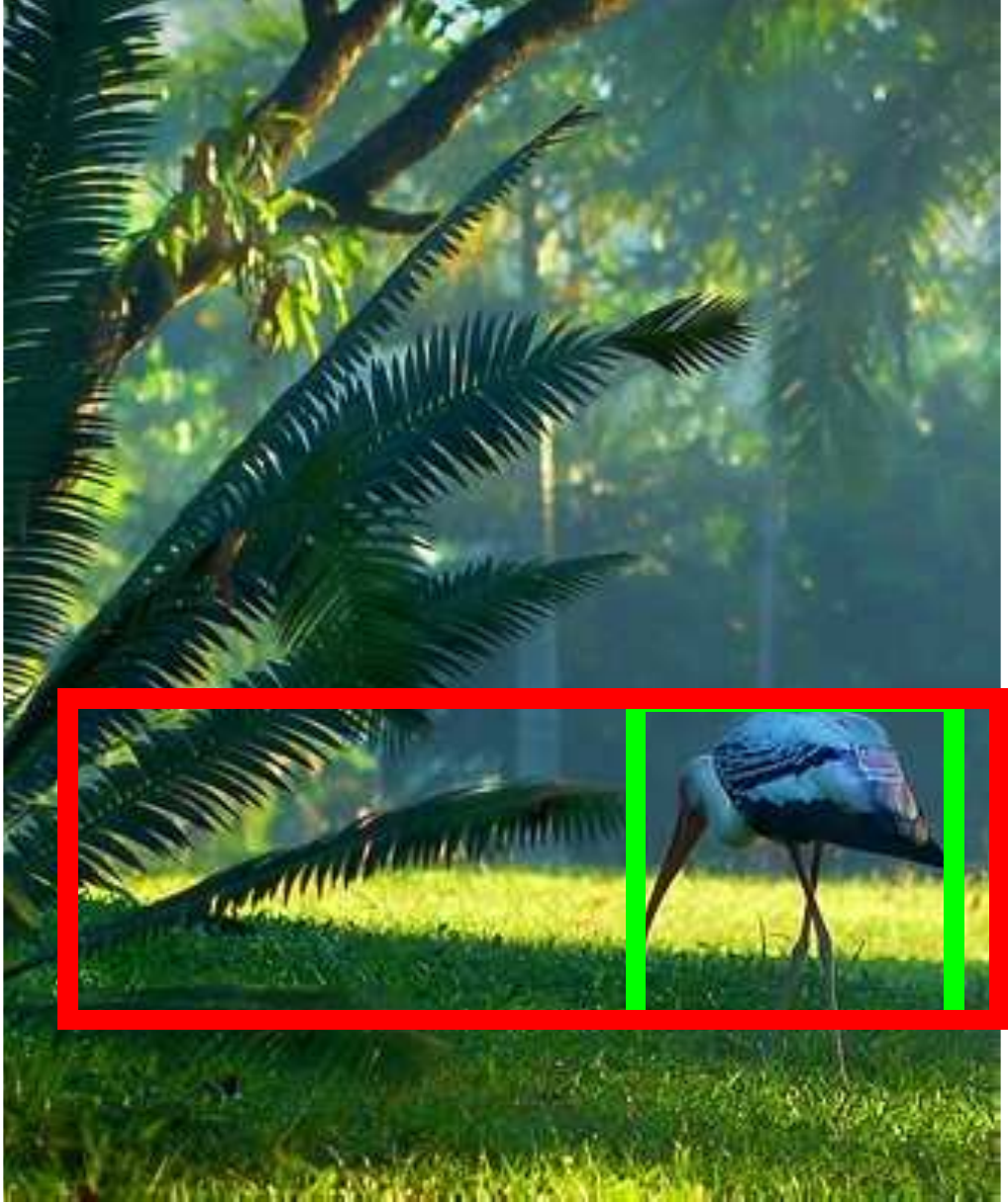}
	} 
	\hspace{0.035 \linewidth}
	\subfloat[HCP+DSD, +OSSH2][HCP+DSD \\ +OSSH2]{\includegraphics[width=0.14   \linewidth,height=0.14  \linewidth]{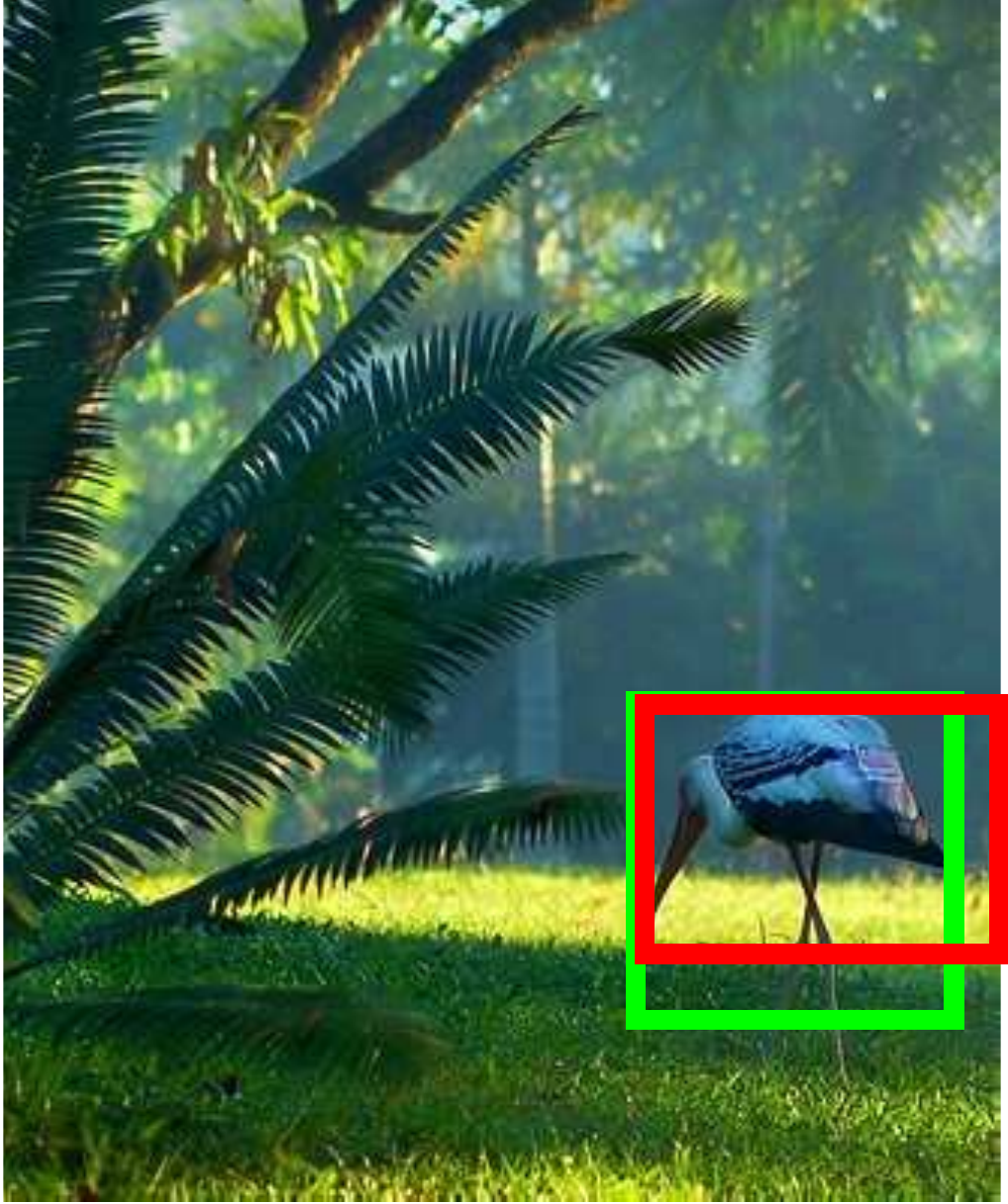}
	}
	\hspace{0.035 \linewidth}
	\subfloat[HCP+DSD, +OSSH3][HCP+DSD \\ +OSSH3]{\includegraphics[width=0.14   \linewidth,height=0.14  \linewidth]{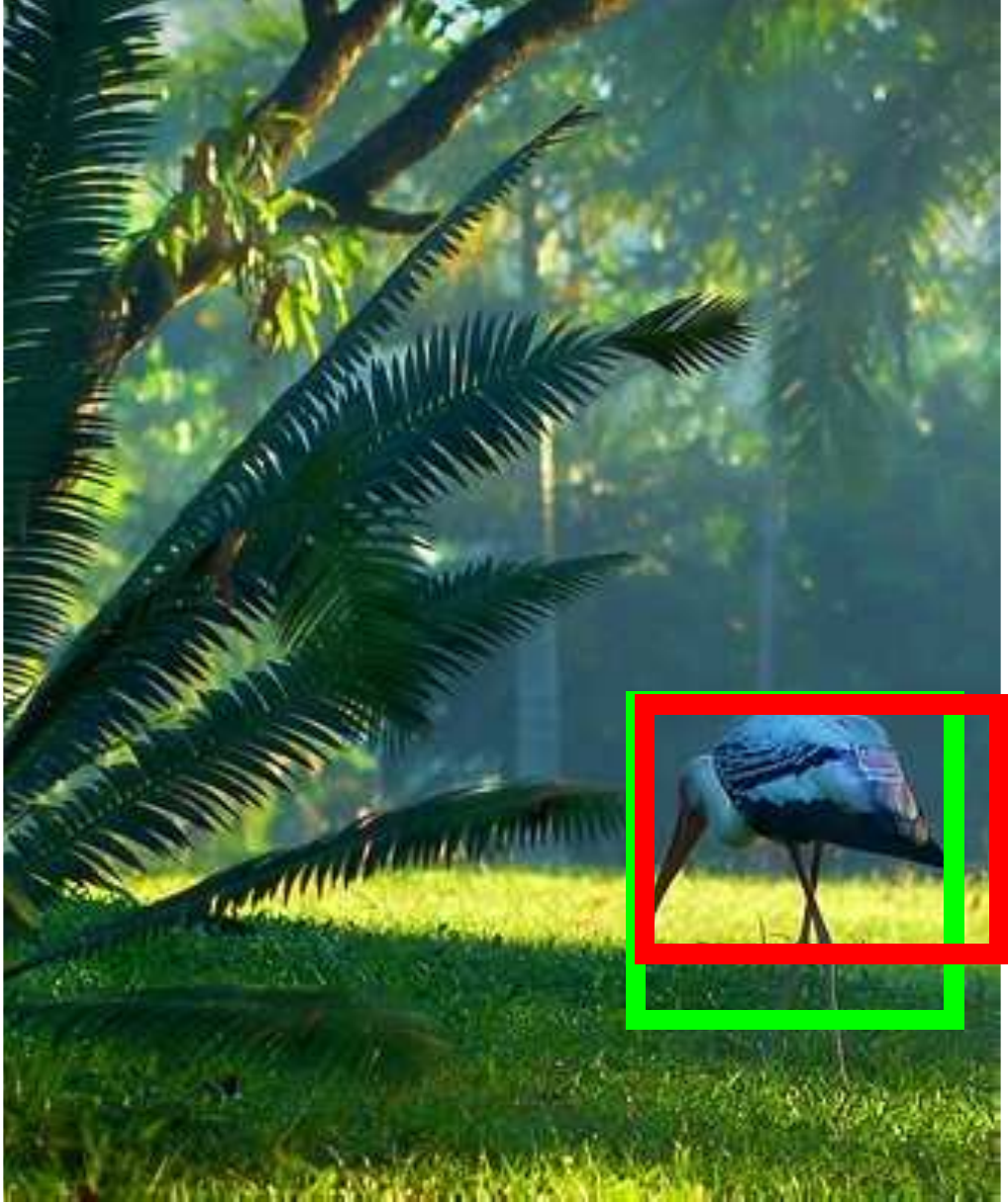}
	}
	\\
	\caption{Qualitative examples of detected objects in different ablation versions of our approach. From the $1$st to the $5$th column: HCP, HCP+DSD, HCP+DSD+OSSH1, HCP+DSD+OSSH2 and HCP+DSD+OSSH3. Green and red bounding boxes represent the ground-truth object bounding boxes and the bounding boxes of the detected objects, respectively. }
	\label{fig:visual}
	\vspace{-4mm}
\end{figure*}

\subsection{Comparison with State-of-The-Arts}
We compare our approach to the state-of-the-art methods. Table \ref{tab:07_trainval_corloc} shows the CorLoc comparison on the PASCAL $2007$ \emph{trainval} set. Our approach achieves the highest result $56.1\%$, compared to all the MIL-based methods (\emph{i.e.},~\cite{cinbis2015multifold,bilen2014weakly,li2016domain}) and the end-to-end WSL network (\emph{i.e.}, \cite{kantorov2016contextlocnet}). Table \ref{tab:07_test} shows the comparison in terms of AP on the PASCAL $2007$ \emph{test} set using the model trained on the PASCAL $2007$ \emph{trainval} set. Our approach achieves $41.7\%$ mAP which also outperforms all the state-of-the-arts, due to the high CorLoc achieved on the corresponding training set (Table \ref{tab:07_trainval_corloc}). With more training data (the PASCAL $2007$ \emph{trainval} set and PASCAL $2012$ \emph{trainval} set), mAP can be further boosted to $43.7\%$ by our approach. Table \ref{tab:12_val} shows the AP comparison on the PASCAL $2012$ \emph{val} set with the state-of-the-art method~\cite{li2016domain}. Both our model and theirs are trained on only the PASCAL $2012$ \emph{train} set. Our approach consistently keeps higher performance, surpassing \cite{li2016domain} by almost $10\%$ in terms of mAP. Table \ref{tab:12_trainval_corloc} gives the comparison between our approach and the state-of-the-art method~\cite{kantorov2016contextlocnet} in terms of CorLoc on the PASCAL $2012$ \emph{trainval} set. The proposed approach significantly outperforms \cite{kantorov2016contextlocnet} by $4\%$ in CorLoc. Table~\ref{tab:12_test} shows AP on the PASCAL $2012$ \emph{test} set of our approach and \cite{kantorov2016contextlocnet} using the models trained on the PASCAL $2012$ \emph{trainval} set. An advantage of $3\%$ on mAP is achieved by our approach. With more training data (the PASCAL $2007$ \emph{trainval} set and PASCAL $2012$ \emph{trainval} set), mAP can be further improved to $39.4\%$ by our method.
\begin{table}
	\small
	\newcommand{\tabincell}[2]{\begin{tabular}{@{}#1@{}}#2\end{tabular}}
	\setlength{\tabcolsep}{3pt}
	\renewcommand{\arraystretch}{1.3}
	\centering
	\caption{Correct localization (CorLoc) (\%) on the PASCAL $2007$ \emph{trainval} set of using relative CNN score improvement and absolute CNN score in OSSH. The comparison is conducted in $3$ cases: performing OSSH in the first $1$, $2$ and $3$ epochs from the $2$nd epoch in training Fast R-CNN.}
	\begin{tabular}{l|cccccccccccccccccccc|c}
		\hline
		Epochs of OSSH & 1 & 2 & 3 \\
		\hline
		absolute CNN score & 48.8 & 52.3 & 53.2 \\
		relative score improvement &  50.2 & 54.9 & 56.1\\
		\hline
	\end{tabular}%
	\label{tab:ablation}%
	\vspace{-0.4cm}
\end{table}%

\subsection{Qualitative Results}
We illustrate examples of detected objects in different ablation versions of our approach in Fig. \ref{fig:visual}. We observe that in some cases the baseline HCP localizes only the key discriminative part of the object, and the localization accuracy can be progressively improved by adding DSD and OSSH to it. Note that in the fifth example which is in the final row of Fig. \ref{fig:visual}, the detected objects by HCP and HCP+DSD are false positive samples which are used as seed positive samples in training the Fast R-CNN detector. By performing OSSH for one epoch, the ground-truth object can be roughly localized, and more epochs of OSSH help precisely select the tight positive proposals, which validates the importance of using relative score improvement in OSSH to avoid the detector being trapped in poor local optima.


\vspace{-0.2cm}
\section{Conclusions}
We proposed a deep self-taught learning approach for weakly supervised object localization. Our approach first acquires effective seed positive object proposals by  examining their response scores to the target class from a classification network, and then mining the spatially concentrated samples via dense subgraph discovery. Then by virtue of online supportive sample harvesting augmented with a new relative CNN score improvement metric, our approach can successfully detect positive samples of improved quality. The experiments demonstrate the superiority of our approach to the state-of-the-art methods. On PASCAL $2007$ and $2012$, the proposed approach consistently outperforms them by an obvious margin in all the evaluation scenarios.
\vspace{-0.1cm}
\section*{Acknowledgments}
Zequn Jie is partially supported by Tencent AI Lab.

The work of Jiashi Feng was partially supported by National University of Singapore startup grant R-263-000-C08-133 and Ministry of Education of Singapore AcRF Tier One grant R-263-000-C21-112.

{\footnotesize
	\bibliographystyle{unsrt}
	\bibliography{egbib}
}

\end{document}